\PassOptionsToPackage{table}{xcolor}
\documentclass[11pt]{article}

\usepackage[preprint]{acl}

\usepackage{times}
\usepackage{latexsym}

\usepackage[T1]{fontenc}

\usepackage[utf8]{inputenc}

\usepackage{microtype}

\usepackage{inconsolata}

\usepackage{graphicx}
\usepackage{placeins}
\usepackage{booktabs}
\usepackage{xcolor}
\definecolor{natureSage}{HTML}{DADDCB}
\definecolor{natureIvory}{HTML}{F5F4EE}
\definecolor{naturePale}{HTML}{E9EEE8}
\definecolor{natureBlueGray}{HTML}{EAF0F2}
\definecolor{natureRule}{HTML}{404844}
\colorlet{SCHead}{natureSage}
\colorlet{SCGroup}{natureIvory}
\colorlet{SCGroupAlt}{naturePale}
\colorlet{SCRule}{natureRule}
\arrayrulecolor{natureRule}
\newcommand{\SCtabletop}{\toprule[0.8pt]}
\newcommand{\SCtablemid}{\midrule[0.45pt]}
\newcommand{\SCtablebottom}{\bottomrule[0.8pt]}
\newcommand{\SCappendixfloats}{%
  \csname @fptop\endcsname=0pt
  \csname @fpsep\endcsname=16pt
  \csname @fpbot\endcsname=0pt plus 1fil
  \csname @dblfptop\endcsname=0pt
  \csname @dblfpsep\endcsname=16pt
  \csname @dblfpbot\endcsname=0pt plus 1fil
}

\usepackage{multirow}
\usepackage{algorithm}
\usepackage{algpseudocode}
\usepackage{amsmath,amssymb,amsthm,mathtools}
\newtheorem{theorem}{Theorem}
\newtheorem{lemma}{Lemma}
\newtheorem{proposition}{Proposition}
\newtheorem{corollary}{Corollary}

\title{Stability-Compensated Threshold Calibration for Selection-Conditioned Risk Control in Large Language Models}

\author{
 \textbf{Zhiyuan Wang}$^{1,\dagger}$,
 \textbf{Yuanchang Ye}$^{2,\dagger}$,
 \textbf{Jiahao Yuan}$^{3}$,
 \textbf{Jinhao Duan}$^{4}$,\\
 \textbf{Xiaoshuang Shi}$^{5}$,
 \textbf{Weijun Xie}$^{6}$,
 \textbf{Ye Wei}$^{6}$,
 \textbf{Kaidi Xu}$^{1,6,*}$,
\\[4pt]
 $^{1}$City University of Hong Kong, $^{2}$National University of Singapore,\\
 $^{3}$East China Normal University, $^{4}$University of North Carolina at Chapel Hill,\\
 $^{5}$University of Electronic Science and Technology of China,\\
 $^{6}$The Hong Kong Institute of AI for Science\\
 \small{
   \textbf{Correspondence:}
   \texttt{kaidixu@cityu.edu.hk}
 }
}
\hypersetup{pdfauthor={}}

\hypersetup{pdftitle={Stability-Compensated Threshold Calibration for Selection-Conditioned Risk Control in Large Language Models}}

\begin{document}
\raggedbottom
\maketitle
\begin{abstract}
Large language models often produce hallucinated answers, motivating selective prediction policies that abstain on uncertain queries.
However, heuristic uncertainty thresholds alone do not provide statistical control of errors among accepted predictions.
Linear Expectation Constraints (LEC) reduce selection-conditioned error rate (SCER) calibration to a nonpositive expected linear-loss constraint, but the associated loss is non-monotonic in the acceptance threshold.
We construct an exact counterexample under i.i.d. sampling demonstrating that LEC's empirical threshold-selection rule can violate the population SCER target.
To address this risk-transfer gap, we introduce Stability-Compensated Linear Expectation Control (SC-LEC), which uses the negative linear risk of a theoretical full-data reference to compensate for threshold-selection instability.
For local expansion beyond a safe anchor, we derive two finite-sample certificates: a stability-based certificate exploiting rank crossings and a direct certificate using the structure of threshold expansion to reduce conservatism.
With a valid population anchor margin, our certificate conditions ensure finite-sample marginal SCER control, and independent margin certification further provides a $1-\delta$ high-probability guarantee for this calibration-averaged control.
Experiments on CommonsenseQA, ARC-Challenge, and TriviaQA show useful retention gains beyond the same certified anchor, while revealing conservative behavior in small-margin regimes.
\end{abstract}

\section{Introduction}
\label{sec:Introduction}

Large language models (LLMs) have demonstrated strong capabilities in question answering, yet remain susceptible to hallucinations and overconfident errors~\citep{huang2025survey}.
Selective prediction offers a practical way to improve reliability by abstaining on uncertain queries rather than answering indiscriminately~\citep{el2010foundations,geifman2019selectivenet}. 
Uncertainty measures such as semantic entropy help determine which predictions to accept~\citep{kuhn2023semantic,duan2024shifting,wang2025word}.
However, a reliable uncertainty ranking does not, by itself, guarantee that the error rate among accepted predictions is statistically controlled.

\begin{figure*}[t]
    \centering
    \includegraphics[width=\textwidth]{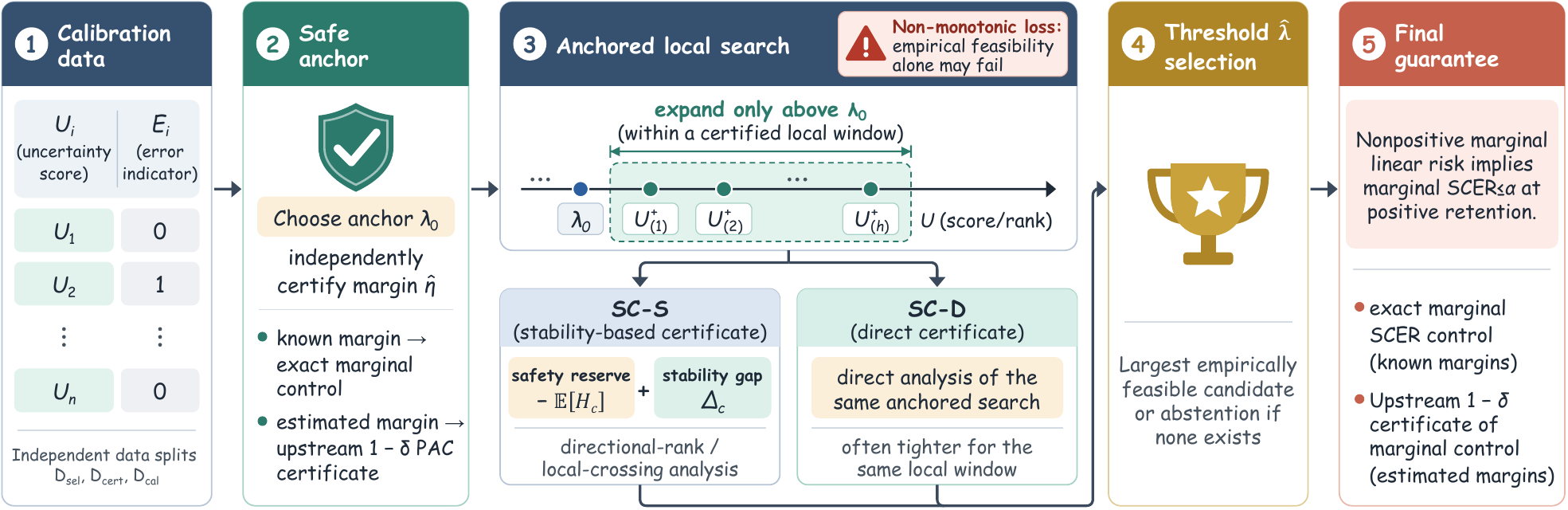}
    \caption{Overview of SC-LEC. A safe anchor supports a one-sided local rank search, certified through stability compensation (SC-S) or direct analysis (SC-D). The rule scans the anchor and up to $h$ subsequent observed scores, selecting the largest empirically feasible candidate or abstaining if none exists. Known valid margins yield finite-sample marginal SCER control at positive retention. Independently certified margins provide an upstream $1-\delta$ certificate of this calibration-averaged guarantee.}
    \label{fig:overview}
\end{figure*}

Our objective is to control the marginal \emph{selection-conditioned error rate} (SCER), defined as the ratio of expected accepted errors to expected selections, on average over both calibration and future observations~\citep{el2010foundations,geifman2017selective}.
Linear Expectation Constraints (LEC)~\citep{wang2026lec} reduce this ratio-form objective to a linear expectation constraint, $\mathbb{E}[(E-\alpha)S]\le 0$, where $E$ and $S$ denote prediction error and selection, respectively, and $\alpha$ is the risk level.
Yet, this linearization alone does not resolve the challenge of adaptive threshold calibration~\citep{angelopoulos2026conformal}. 
The induced loss is non-monotonic: accepting a correct prediction decreases the cumulative linear loss by $\alpha$, whereas accepting an incorrect prediction increases it by $1-\alpha$. 
Consequently, standard monotonic calibration arguments do not directly justify selecting the largest empirically feasible threshold. 
An exact i.i.d. counterexample shows that the original LEC complete-scan rule can violate the population SCER target (Figure~\ref{fig:lec-exact-counterexample}). 
Moreover, a positive expected linear-risk slack $\beta$ yields only $\mathrm{SCER}\le\alpha+\beta/r$, where $r$ denotes marginal retention, making exact control particularly challenging when retention is small.

We propose Stability-Compensated Linear Expectation Control (\textbf{SC-LEC}), which establishes finite-sample guarantees for anchored non-monotonic threshold calibration. 
As shown in Figure~\ref{fig:overview}, SC-LEC compensates for threshold-selection instability using the negative linear risk of a theoretical full-data reference.
Building on the stability-transfer perspective of~\citet{angelopoulos2026conformal}, we decompose the deployed linear risk into a safety reserve and a signed stability gap.
When the reserve dominates a valid upper bound on this gap, the resulting linear risk remains nonpositive, recovering the desired marginal SCER control.
To make this principle operational, we introduce an anchored local-search procedure and derive two complementary finite-sample certificates: a stability-based certificate that exploits directional rank crossings and a direct certificate that uses the structure of one-dimensional threshold expansion to reduce conservatism. 
Independent margin certification further enables the certificates to be constructed from samples without requiring a known population margin. 

These results address both the risk-transfer gap and its practical certification: 
Firstly, we identify a failure of empirical non-monotonic threshold calibration through an exact population counterexample and specialize reserve--stability analysis to the selection-conditioned linear-risk target. 
Moreover, we develop structure-aware finite-sample certificates for anchored threshold search, including directional and local-crossing stability bounds and a tighter direct analysis that reduces certification conservatism.
An executable calibration procedure combines independent margin certification with automatic buffer and window selection. 
With valid population margins, the resulting designs achieve finite-sample marginal SCER control, while independently certified margins provide a $1-\delta$ high-probability guarantee of this calibration-averaged control. 
Experiments across three QA benchmarks show expansion beyond a common safe anchor and limited certification in small-margin regimes.

\section{Related Work}
\label{sec:Related Work}

\noindent \textbf{Selective Prediction and Risk Calibration.} Selective prediction improves reliability by allowing models to abstain on uncertain inputs, trading prediction coverage for reduced error~\citep{el2010foundations,geifman2017selective,geifman2019selectivenet}.
For language generation, semantic uncertainty estimates uncertainty by clustering equivalent outputs~\citep{kuhn2023semantic,farquhar2024semantic,lin2024generating}. Conformal approaches also study uncertainty and selection in language models~\citep{wang2024conu,wang2025sconu}. 
Yet, uncertainty quantification does not ensure statistical reliability, motivating calibration methods with explicit risk guarantees. 
Conformal prediction offers distribution-free coverage guarantees~\citep{angelopoulos2023conformal,wang2024conu,wang2025sconu}, while conformal risk control (CRC) extends this perspective to monotonic expected losses~\citep{angelopoulos2024conformal}.
Learn then Test further enables high-probability risk control through statistical testing~\citep{bates2021distribution,angelopoulos2025learn}.
COIN studies calibrated abstention for question answering~\citep{wang2026coin}, while Trust or Escalate studies selective LLM judging with human-agreement guarantees~\citep{jung2025trust}.
These studies motivate statistically rigorous selection, but differ in their risk targets and calibration procedures.

\noindent \textbf{Linear and Non-Monotonic Calibration.} 
LEC frames marginal selection-conditioned error control as a linear expectation constraint~\citep{wang2026lec}, providing a foundation for selective prediction and routing~\citep{jia2026balancerag}. 
Although this reduction is exact, the resulting loss is non-monotonic in the acceptance threshold, and empirical feasibility alone does not justify population risk control under adaptive selection. 
We demonstrate this limitation via an exact counterexample to LEC's unrestricted complete-scan rule in Appendix~\ref{app:lec-counterexample}. 
Related approaches address non-monotonicity through different mechanisms: monotonized calibration provides asymptotic control~\citep{lin2026asymptotic}, whereas finite-grid analyses establish sample-dependent risk bounds~\citep{aldirawi2026conformal}.
SC-LEC studies calibration-generated thresholds through anchored local search with explicit risk compensation for finite-sample marginal SCER control.
Further connections are discussed in Appendix~\ref{app:additional-related-work}.

\noindent \textbf{Stability of Threshold Calibration.} 
Our work is closely related to the framework of \citet{angelopoulos2026conformal}, which transfers expected loss from a full-data reference to a calibration-fitted rule for non-monotonic risk control.
This analysis also develops rank-crossing bounds for selective classification, which our analysis builds on.
Nonetheless, for the shifted selective loss $\widetilde{\ell}_\alpha=g_\alpha+\alpha$, an expected-loss bound with additive slack $\beta$ yields $\operatorname{SCER}
\le \alpha+\frac{\beta}{r}$, where $r=\mathbb{E}[S]>0$, from the linear-risk identity alone, making the slack sensitive to marginal retention. 
SC-LEC instead targets the nonpositive linear-risk endpoint by explicitly compensating threshold-selection instability with a sufficient safety reserve.
Beyond this principle, we derive directional and local-crossing bounds, develop a tighter direct certificate for anchored expansion, and independently certify the required population margin.
While the prior work explores bootstrap-based stability estimation, it does not establish a general finite-sample guarantee for that estimator. Our explicit certificates avoid relying on such estimates.

\section{Method}
\label{sec:method}
We derive a reserve condition and two anchored-search certificates for marginal SCER control, then certify anchor margins independently.

\subsection{Problem Setup and Linear Target}
\label{sec:problem}
Let $W_i=(U_i,E_i)$, where the components are a fixed predictor's uncertainty score and error indicator $E_i\in\{0,1\}$. Lower scores indicate greater confidence. Assume $W_{1:N}$ is exchangeable, with $N=n+1$:
$D_{1:n}$ calibrates the threshold and $W_N$ is the test observation.
Define the selection and accepted error by $S_i(\lambda)=\mathbf1\{U_i\le\lambda\}$ and $Z_i(\lambda)=E_iS_i(\lambda)$, respectively. 
For target $\alpha\!\in\!(0,1)$, the linear loss is
\begin{equation}
g_i(\lambda)=g_\alpha(W_i;\lambda)
=Z_i(\lambda)-\alpha S_i(\lambda).
\label{eq:selection}
\end{equation}
For a calibrated threshold $\widehat\lambda=A(D_{1:n})$, we define
\begin{equation}
\begin{aligned}
\operatorname{SCER}(\widehat\lambda)
&=\frac{\mathbb E[Z_N(\widehat\lambda)]}{\mathbb E[S_N(\widehat\lambda)]},\\
\mathcal R_n(A)&=\mathbb E[g_N(\widehat\lambda)].
\end{aligned}
\label{eq:scer}
\end{equation}
Expectations average calibration and test randomness. SCER conditions on selection in this joint experiment, not on each realized threshold, and is undefined at zero retention. For
$r_n(A)=\mathbb E[S_N(\widehat\lambda)]>0$,
\[
\operatorname{SCER}(\widehat\lambda)=\alpha+\mathcal R_n(A)/r_n(A).
\]
Hence, LEC's reduction~\citep{wang2026lec} gives
\begin{equation}
\operatorname{SCER}(\widehat\lambda)\le\alpha
\; \Longleftrightarrow \; \mathcal R_n(A)\le0.
\label{eq:linear-equivalence}
\end{equation}
The loss lies in $[-\alpha,1-\alpha]$ but is non-monotonic: correct
and incorrect newly accepted predictions contribute $-\alpha$ and
$1-\alpha$, respectively. We prove the equivalence in Appendix~\ref{app:proof-linear-scer}. 

\subsection{Margin--Stability Decomposition}
\label{sec:buffered-calibration}
For a sample $D_m$, let $T_m(\lambda)=\sum_{i=1}^m g_i(\lambda)$.
The observed scores form the candidate set
$\mathcal T(D_m)=\{U_1,\ldots,U_m\}$. A fixed buffer $c\ge0$
defines its feasible subset
\[
\mathcal F_c(D_m)=\{\lambda\in\mathcal T(D_m):T_m(\lambda)\le-c\}.
\]
The permutation-invariant calibration rule selects
\begin{equation}
A_c(D_m)=
\begin{cases}
\max\mathcal F_c(D_m),&\mathcal F_c(D_m)\ne\varnothing,\\
\bot,&\text{otherwise}.
\end{cases}
\label{eq:empirical-linear-sum}
\end{equation}
Here, $\bot$ represents full abstention, with
$S_i(\bot)=Z_i(\bot)=g_i(\bot)=0$.
$c=1$ recovers LEC's
empirical constraint~\citep{wang2026lec}.

Let $D_{-i}$ denote $D_N$ with $W_i$ removed. We compare the full-data and leave-one-out thresholds:
\begin{equation}
\widehat\lambda^\star=A_c(D_N) \text{ and }
\widehat\lambda_{-i}=A_c(D_{-i}).
\label{eq:oracle-threshold}
\end{equation}
The full-data rule is an analytical reference; deployment uses calibration data alone. Its feasibility event and reserve are
\begin{equation}
F_c=\{\mathcal F_c(D_N)\ne\varnothing\} \text{ and }
H_c=-T_N(\widehat\lambda^\star)/N.
\label{eq:Fc}
\end{equation}
Lemma~\ref{lem:oracle-risk} in Appendix~\ref{app:proof-oracle-risk} gives $\mathbb E[H_c]\ge(c/N)\Pr(F_c)$. The leave-one-out gap is
\begin{equation}
\Delta_c=\mathbb E\!\left[\frac1N\sum_{i=1}^N
\bigl(g_i(\widehat\lambda_{-i})-g_i(\widehat\lambda^\star)\bigr)\right].
\label{eq:exact-stability-gap}
\end{equation}
\begin{theorem}[Margin--stability decomposition]
\label{thm:decomposition}
Under exchangeability and permutation invariance,
\begin{equation}
\mathcal R_n(A_c)=-\mathbb E[H_c]+\Delta_c.
\label{eq:exact-decomposition}
\end{equation}
Consequently, any upper bound $\Delta_c\le\beta_c$ gives
\begin{equation}
\mathcal R_n(A_c)\le\beta_c-\mathbb E[H_c].
\label{eq:decomposition-upper}
\end{equation}
\end{theorem}
The signed gap can be positive or negative. A reserve dominating its upper bound ensures nonpositive risk.
In particular,
\[
\Delta_c\le\beta_c\le(c/N)\Pr(F_c)
\;\Longrightarrow\;\mathcal R_n(A_c)\le0.
\]
The feasibility factor accounts for zero reserve when the reference abstains. 
We prove the identity and shift
invariance of the stability gap in Appendix~\ref{app:proof-decomposition}.

\subsection{Structure-Aware Stability and SCER Control}
\label{sec:stability}
Assume almost surely distinct scores~\citep{angelopoulos2026conformal}.\footnote{A fixed independent
continuous key resolves ties by a strict lexicographic order. See details in Appendix~\ref{app:server-scores}.}
Let $r_N(\lambda)=\sum_{j=1}^N S_j(\lambda)$, $r_N(\bot)=0$,
$r^\star=r_N(\widehat\lambda^\star)$, and
$r_i=r_N(\widehat\lambda_{-i})$. Upward and downward rank
displacements are
\begin{equation}
\begin{aligned}
K_c^+&=\max_i(r_i-r^\star)_+,\\
K_c^-&=\max_i(r^\star-r_i)_+,
\end{aligned}
\label{eq:directional-K}
\end{equation}
with $K_c=\max\{K_c^+,K_c^-\}$ and $(x)_+$ representing $\max\{x,0\}$.
\begin{proposition}[Directional rank stability]
\label{prop:rank-stability}
Under the preceding assumptions,
\begin{equation}
\begin{aligned}
\Delta_c\le\beta_c&:=
\frac{(1-\alpha)\mathbb E[K_c^+]+\alpha\mathbb E[K_c^-]}{N}\\
&\le\frac{\mathbb E[K_c]}{N}.
\end{aligned}
\label{eq:beta-rank-directional}
\end{equation}
\end{proposition}
An upward acceptance crossing costs at most $1-\alpha$, and a downward
crossing costs at most $\alpha$. Keeping direction sharpens the generic
absolute bound $2\max\{\alpha,1-\alpha\}\mathbb E[K_c]/N$.
Appendix~\ref{app:proof-rank-stability} gives the counting argument.
\begin{samepage}
\begin{theorem}[Stability-compensated SCER control]
\label{thm:scer-control}
Under exchangeability, permutation invariance, and almost surely
distinct scores,
\begin{equation}
\mathcal R_n(A_c)\le\beta_c-\frac cN\Pr(F_c).
\end{equation}
If
\begin{equation}
(1-\alpha)\mathbb E[K_c^+]+\alpha\mathbb E[K_c^-]\le c\Pr(F_c),
\label{eq:main-condition}
\end{equation}
then $\mathcal R_n(A_c)\le0$ and SCER is at most $\alpha$
whenever retention is positive.
\end{theorem}
\end{samepage}
This unrestricted-search result requires population stability and feasibility bounds, but neither i.i.d. sampling nor a safe anchor. Anchored search makes these bounds explicit. Appendix~\ref{app:proof-scer-control} gives the proof; further rank bounds and a decision-rule extension appear in
Appendix~\ref{app:supplementary-rank}.

\subsection{Explicit Anchored Certificates}
\label{sec:explicit-instantiation}
Now assume $W_{1:N}\stackrel{\rm i.i.d.}{\sim}P$ with almost surely
distinct scores. A \emph{prespecified design} fixes $\lambda_0,\eta,c,h$
independently of final calibration, with $c\ge0$ and integer $h\ge0$.
The safe anchor satisfies
\begin{equation}
\mu_0:=\mathbb E[g_\alpha(W;\lambda_0)]\le-\eta,
\; 0<\eta<\alpha.
\label{eq:safe-core}
\end{equation}
Writing $p_0=\Pr(U\le\lambda_0)$, the anchor condition becomes
\[
p_0[\alpha-\Pr(E=1\mid U\le\lambda_0)]\ge\eta.
\]
Thus reserve combines retention and error headroom: even a highly
accurate anchor can have little margin if it accepts very little.
No perfect uncertainty ranking is assumed.

Let $r_0(D_m)=\#\{i:U_i\le\lambda_0\}$ and $U^+_{(j)}$ be the
$j$th score above the anchor. The radius $h$ counts observed ranks:
with $h_m=\min\{h,m-r_0(D_m)\}$, the candidate set is
\[
\mathcal T_h(D_m)=\{\lambda_0\}\cup\{U^+_{(j)}:1\le j\le h_m\}.
\]
Its feasible subset is
\[
\mathcal F_{c,h}(D_m)=\{\lambda\in\mathcal T_h(D_m):T_m(\lambda)\le-c\}.
\]
The anchored rule is
\begin{equation}
A_{c,h}(D_m)=
\begin{cases}
\max\mathcal F_{c,h}(D_m),&\text{if nonempty},\\
\bot,&\text{otherwise}.
\end{cases}
\label{eq:local-candidates}
\end{equation}
Every non-abstaining output accepts the entire anchor. Non-monotonic feasibility requires scanning all candidates, even when the anchor is infeasible. Define $F_{c,h},H_{c,h},\Delta_{c,h}$
by substituting this rule and feasible set in the preceding definitions, and Theorem~\ref{thm:decomposition} still applies.

For binary KL divergence $\operatorname{kl}(x\Vert y)$, set
\begin{equation}
q_{m,b}=\exp\!\left[-m\operatorname{kl}
\left(\alpha-\frac bm\,\middle\Vert\,\alpha-\eta\right)\right]
\label{eq:anchor-kl-tail}
\end{equation}
when $0\le b<m\eta$, and $q_{m,b}=1$ when $b\ge m\eta$.
Lemma~\ref{lem:anchor-feasible} bounds
$\Pr\{T_m(\lambda_0)>-b\}$ by $q_{m,b}$ using only the range and mean, as proved in Appendix~\ref{app:proof-anchor-feasible}. Write
$\widetilde q_{N,c}=q_{N,c+\alpha}$,
$p_{N,c}=1-q_{N,c}$, and $M_\alpha=\max\{\alpha,1-\alpha\}$.
\begin{theorem}[Explicit stability certificate: SC-S]
\label{thm:explicit-scer}
Under this subsection's assumptions, if $c+\alpha<N\eta$ and
$0\le h\le n$, define
\begin{equation}
\gamma_{N,c,h}=M_\alpha[h+(N-h)\widetilde q_{N,c}].
\label{eq:gamma-tight}
\end{equation}
Then
\begin{equation}
\mathcal R_n(A_{c,h})\le\frac{\gamma_{N,c,h}-c\,p_{N,c}}N.
\label{eq:explicit-risk-bound}
\end{equation}
In particular,
\begin{equation}
\gamma_{N,c,h}\le c\,p_{N,c}
\label{eq:explicit-condition-tight}
\end{equation}
ensures SCER at most $\alpha$ whenever retention is positive.
\end{theorem}
On $\{T_N(\lambda_0)\le-c-\alpha\}$, the anchor remains feasible in every
deleted sample, and only the first $h$ above-anchor points can change
their own acceptance. Appendix~\ref{app:proof-explicit-scer} bounds
the complementary event and combines crossings with reserve.

SC-S makes the full-data/LOO reserve--stability principle explicit through local crossing bounds. SC-D exploits the same one-dimensional rule to bound anchor and expansion contributions directly. Under the default KL rules, a shared anchor, margin, sample size, and score order at $\alpha\le1/3$ imply that SC-D accepts every point accepted by SC-S. Appendix~\ref{app:default-design-ordering} proves this inclusion. It does not order realized SCER or arbitrary designs. Both certificates allow positive buffers.
\label{sec:direct-certificate}
\pagebreak[3]
\begin{theorem}[Direct anchor-expansion bound: SC-D]
\label{thm:direct-anchor}
Under the i.i.d., prespecified-design, and distinct-score assumptions
above, if $0\le c<n\eta$ and $0\le h\le n$, then
\begin{equation}
\mathcal R_n(A_{c,h})\le-\eta(1-q_{n,c})+\frac{(1-\alpha)h}{N}.
\label{eq:direct-anchor-risk}
\end{equation}
Consequently, if
\begin{equation}
(1-\alpha)h\le N\eta(1-q_{n,c}),
\label{eq:direct-anchor-condition}
\end{equation}
SCER is at most $\alpha$ whenever retention is positive.
\end{theorem}
Rank symmetry bounds fresh-test acceptance beyond the anchor by $h/N$ (Appendix~\ref{app:proof-direct-anchor}). Negative anchor risk offsets worst-case expansion loss. Corollary~\ref{cor:retention}
gives positive retention for both: SC-S implies $c<n\eta$ since
$\eta<\alpha$. A passing anchor may fail SC-S, which also accounts for loss of simultaneous deleted-sample feasibility. Certificate unavailability does not establish unsafety.

\noindent \textbf{Certification tightness and retention.}
For fixed $\lambda_0,c,D_n$ and score/tie order, increasing $h$ nests
the feasible sets and cannot decrease the output or its acceptance,
with $\bot$ ordered first. Appendix~\ref{app:nested-windows} proves this nesting property.
A sharper radius certificate can therefore improve retention at the
same marginal target. The argument does not order risk or designs
with different buffers.

\subsection{Data-Driven Anchor Certification}
\label{sec:anchor-certification}
Fix the predictor, score, sizes $m,n\ge1$, target $\alpha$, budget $\delta\in(0,1)$, and selection policy. Split i.i.d. data into independent roles: $D_{\rm sel}$ constructs a nonempty grid $\{\lambda_j\}_{j=1}^L$, $D_{\rm cert}$ certifies margins, $D_{\rm cal}$ selects the final threshold, and test data evaluate it. Certification conditions on the grid~\citep{angelopoulos2025learn}. An external grid needs no $D_{\rm sel}$. Average over \emph{all} $m$ certification points:
\[
\overline Y_j=\frac1m\sum_{W\in D_{\rm cert}}
[g_\alpha(W;\lambda_j)+\alpha].
\]
Invert the bounded-variable KL tail to obtain
\begin{equation}
\begin{aligned}
u_j&=\sup\bigl\{u\in[\overline Y_j,1]:\\
&\qquad m\operatorname{kl}(\overline Y_j\Vert u)\le\log(L/\delta)\bigr\}.
\end{aligned}
\label{eq:certification-average}
\end{equation}
The binary KL divergence is defined by continuity at the endpoints. With
$\mu_j=\mathbb E[g_\alpha(W;\lambda_j)]$, Chernoff and a union bound give
\begin{equation}
\Pr\{\mu_j\le u_j-\alpha\text{ for all }j\}\ge1-\delta.
\label{eq:simultaneous-anchor-certificate}
\end{equation}
Select a passing candidate $u_{j^\star}<\alpha$ and set
\begin{equation}
\lambda_0=\lambda_{j^\star},\;
\widehat\eta=\alpha-u_{j^\star}\in(0,\alpha).
\label{eq:certified-anchor-margin}
\end{equation}
The simultaneous bound in Eq.~\eqref{eq:simultaneous-anchor-certificate} permits anchor selection using certification results, as proved in Appendix~\ref{app:proof-margin-certification}. The margin policy
maximizes $\widehat\eta$, while the alternative coverage policy takes the
largest passing anchor. If none passes, abstain. The certified quantity
is unconditional linear risk, including the anchor's retention factor.

Before examining $D_{\rm cal}$, SC-D sets $c=0$ and chooses the largest
integer $h\le n$ satisfying Eq.~\eqref{eq:direct-anchor-condition} with
$\eta=\widehat\eta$. SC-S searches a fixed finite buffer grid for the
largest certified radius, breaking ties by smaller $c$, and unavailable
designs follow a fixed fallback (abstention in our experiments).
Appendix~\ref{app:automatic-parameters} gives closed forms and
conservative numerical evaluation.

Let $H=(D_{\rm sel},D_{\rm cert})$ and
$\widehat\lambda=A_H(D_{\rm cal})$. For a fresh $W$, define
\begin{equation}
\begin{aligned}
\mathcal R(H)&=\mathbb E[g_\alpha(W;\widehat\lambda)\mid H],\\
r(H)&=\mathbb E[S(W;\widehat\lambda)\mid H].
\end{aligned}
\label{eq:upstream-conditional-risk}
\end{equation}
Here $S(W;\lambda)=\mathbf1\{U\le\lambda\}$ and $S(W;\bot)=0$.
Conditioning on $H$ fixes the anchor, margin, buffer, and radius, while expectations average final calibration and fresh testing.
\begin{theorem}[Certified marginal SCER]
\label{thm:certified-marginal-scer}
Under the independent i.i.d. split and distinct-score assumption,
the simultaneous certificate and either valid parameter rule give
\begin{equation}
\Pr_H\{\mathcal R(H)\le0\}\ge1-\delta.
\label{eq:pac-marginal-linear-risk}
\end{equation}
Except on an upstream event of probability at most $\delta$, every
passing design has $\operatorname{SCER}_H\le\alpha$ and
\begin{equation}
r(H)\ge\frac{\widehat\eta}{\alpha}(1-q_{n,c})>0.
\label{eq:certified-retention-floor}
\end{equation}
\end{theorem}
Here, $\operatorname{SCER}_H=\alpha+\mathcal R(H)/r(H)$.
The probability is over $H$; $\mathcal R(H)$ still averages final calibration and testing. This does not imply per-threshold guarantees or confidence conditional on passing.
With estimated margins, averaging over upstream failures gives only
$\mathbb E[g_\alpha]\le(1-\alpha)\delta$ and, at positive overall retention,
\[
\operatorname{SCER}_{\rm all}\le\alpha+
\frac{(1-\alpha)\delta}{\mathbb E[S]}.
\]
Failure-triggered abstention does not remove this slack, as shown in
Appendices~\ref{app:proof-certified-marginal-scer}--\ref{app:certification-implementation}.
The tails $q_{m,b}$ are already integrated into risk and add no
confidence penalty to $\delta$.

\subsection{Implementation}
\label{sec:implementation}
Sort calibration scores, compute prefix losses, and scan every allowed
candidate. The anchored scan checks at most $h+1$ candidates, and sorting
dominates the $O(n\log n)$ cost. Algorithm~\ref{alg:sclec} and
Appendix~\ref{app:automatic-parameters} specify the procedure and
parameters. Prompts, score calculations, and a worked scan appear in
Appendices~\ref{app:prompts}--\ref{app:calibration-example}.

\begin{figure*}[!t]
\centering
\includegraphics[width=\textwidth]{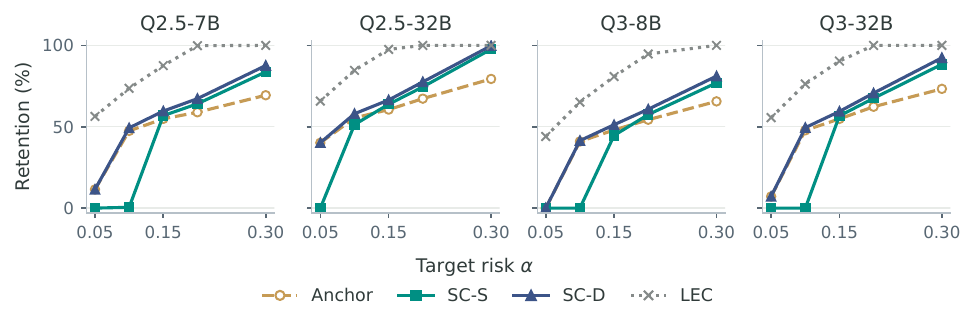}
\caption{Mean CommonsenseQA retention across five target risks and all 100 paired repetitions, including failures. Anchor, SC-S, and SC-D use the margin/KL design; marker shapes and line styles identify methods. LEC is an empirical reference.}
\label{fig:csqa-retention}
\end{figure*}

\begin{table*}[!t]
\centering
\begingroup
\small
\begin{tabular}{llrrrr}
\SCtabletop
\rowcolor{SCHead}
 & & Empirical & \multicolumn{3}{c}{Certified anchored} \\
\cmidrule(lr){3-3}\cmidrule(lr){4-6}
\rowcolor{SCHead}
Dataset & Model & LEC & Anchor & SC-S & SC-D \\
\SCtablemid
\rowcolor{SCGroup}
 & Qwen2.5-7B & 99.77\,/\,19.43 & 58.88\,/\,5.59 & 64.06\,/\,7.09 & \textbf{67.23}\,/\,8.03 \\
\rowcolor{SCGroup}
 & Qwen2.5-32B & 99.96\,/\,16.00 & 67.25\,/\,5.38 & 74.46\,/\,7.00 & \textbf{77.52}\,/\,7.76 \\
\rowcolor{SCGroup}
 & Qwen3-8B & 94.67\,/\,20.04 & 54.30\,/\,7.12 & 57.45\,/\,7.93 & \textbf{60.73}\,/\,8.77 \\
\rowcolor{SCGroup}
\multirow{-4}{*}{CommonsenseQA} & Qwen3-32B & 99.94\,/\,18.85 & 62.27\,/\,6.43 & 67.47\,/\,7.64 & \textbf{70.64}\,/\,8.49 \\
\cmidrule(lr){1-6}
 & Qwen2.5-7B & 99.83\,/\,10.42 & 80.56\,/\,4.13 & 87.13\,/\,5.95 & \textbf{92.84}\,/\,7.83 \\
 & Qwen2.5-32B & 99.86\,/\,4.68 & 93.29\,/\,2.56 & 99.83\,/\,4.67 & \textbf{99.86}\,/\,4.68 \\
 & Qwen3-8B & 99.84\,/\,9.31 & 84.50\,/\,3.63 & 92.73\,/\,6.20 & \textbf{97.72}\,/\,8.25 \\
\multirow{-4}{*}{ARC-Challenge} & Qwen3-32B & 99.84\,/\,4.73 & 91.27\,/\,2.00 & 99.81\,/\,4.72 & \textbf{99.84}\,/\,4.73 \\
\cmidrule(lr){1-6}
\rowcolor{SCGroup}
 & Qwen2.5-7B & 34.69\,/\,19.89 & \textbf{0.40}\,/\,12.19 & 0.00\,/\,-- & \textbf{0.40}\,/\,12.19 \\
\rowcolor{SCGroup}
 & Qwen2.5-32B & 63.79\,/\,20.23 & 37.15\,/\,9.86 & 0.00\,/\,-- & \textbf{38.96}\,/\,10.35 \\
\rowcolor{SCGroup}
 & Qwen3-8B & 37.04\,/\,20.18 & \textbf{5.78}\,/\,10.48 & 0.00\,/\,-- & \textbf{5.78}\,/\,10.48 \\
\rowcolor{SCGroup}
\multirow{-4}{*}{TriviaQA} & Qwen3-32B & 39.21\,/\,19.92 & 23.19\,/\,9.10 & 0.00\,/\,-- & \textbf{23.26}\,/\,9.14 \\
\SCtablebottom
\end{tabular}
\endgroup

\caption{Retention / empirical SCER (\%) at $\alpha=0.20$, aggregated over all 100 repetitions, including failed and unavailable designs. SC methods use the margin/KL design. A dash denotes undefined SCER at zero retention. SC-S is unavailable on TriviaQA at this target. Boldface indicates the highest retention among Anchor, SC-S, and SC-D, including ties at the displayed precision.}
\label{tab:csqa-main}
\label{tab:server-transfer}
\label{tab:cross-dataset}
\end{table*}

\section{Experiments}
\label{sec:experiments}
We examine expansion beyond a shared safe anchor, relating its extent to reserve and threshold-selection instability.

\subsection{Setup}
\label{sec:experiment-setup}
We use CommonsenseQA~\citep{talmor2019commonsenseqa},
ARC-Challenge~\citep{clark2018think}, and closed-book
TriviaQA~\citep{joshi2017triviaqa}, with Qwen2.5-7B/32B-Instruct
and Qwen3-8B/32B~\citep{qwen2025qwen25,yang2025qwen3} models under one zero-shot protocol. 
Uncertainty score is measured by the normalized option entropy for multiple choice and mean answer-token
NLL for TriviaQA. Each of 100 paired repetitions utilizes
10/35/25/30\% selection/certification/calibration/test splits.
We fix $\alpha\in\{0.05,0.10,0.15,0.20,0.30\}$, $\delta=0.05$ per
model/score/target, and 20 selection-derived candidates. The main
margin/KL policy maximizes certified margin before final calibration;
coverage anchors and Hoeffding bounds are prespecified ablations.

\begin{figure*}[t]
\centering
\includegraphics[width=\textwidth]{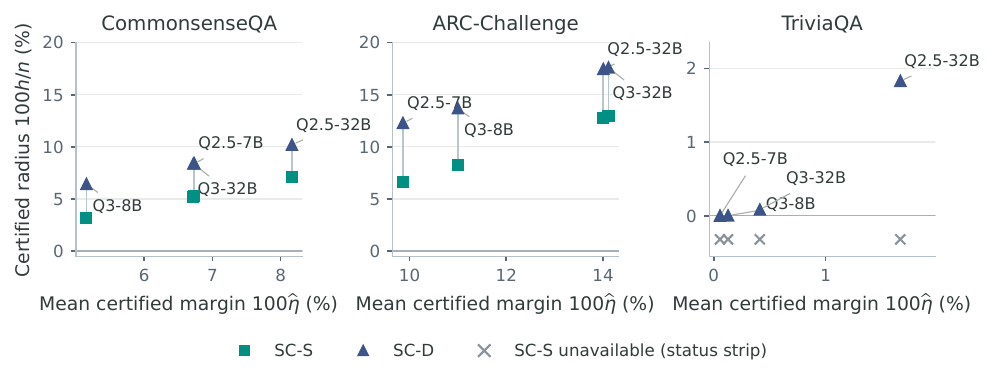}
\caption{Certified margin and search radius at $\alpha=0.20$ (margin/KL design). Labels identify Qwen family and size. Points average available repetitions within each branch; segments connect branches with identical upstream cohorts. Gray crosses in the status strip mark unavailable SC-S designs.}
\label{fig:certificate-geometry}
\end{figure*}

\begin{figure*}[!t]
\centering
\includegraphics[width=\textwidth]{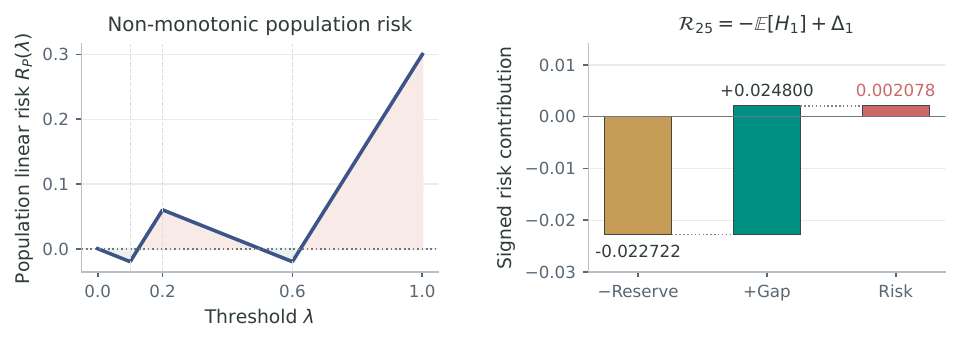}
\caption{Exact four-band i.i.d. counterexample with $n=25$, $c=1$, and $\alpha=0.20$. Left: population linear risk by threshold. Right: the signed stability gap exceeds the reserve, giving $\mathcal R_{25}\approx0.002078>0$ and marginal SCER $\approx0.2080>0.20$. Appendix~\ref{app:lec-counterexample} derives these quantities by exact enumeration.}
\label{fig:lec-exact-counterexample}
\end{figure*}

We compare \textbf{Anchor}, \textbf{SC-S}, and \textbf{SC-D} using the
same certified anchor, and complete-scan \textbf{LEC} ($c=1$).
Anchor uses $m$ certification labels, LEC $n$ calibration labels, and SC-S/D $m+n$ labels in independent roles, so label budgets differ. Retention is averaged over repetitions; empirical SCER divides pooled accepted errors by pooled accepted predictions.
Failed or unavailable designs abstain and remain included; SCER is undefined at zero retention. These finite-pool summaries do not measure population PAC failure. Appendices~\ref{app:server-protocol}--\ref{app:server-results}
give scores, label budgets, all targets, and ablations. LTT-cal and LTT-pool exceed SC-D retention in 51 of 60 primary cells (three ties, six lower). LTT tests the fitted policy with high probability; SC-LEC certifies calibration-averaged marginal control for a local rank rule generated by final calibration, with confidence over upstream margin certification. Appendix~\ref{app:additional-related-work} details guarantees and label use~\citep{angelopoulos2025learn}.

\subsection{Certified Expansion on Real Tasks}
\label{sec:experiment-main-results}
\label{sec:experiment-transfer}
At $\alpha=0.20$, CommonsenseQA Qwen2.5-32B expands from 67.25\%
Anchor retention to 74.46\% for SC-S and 77.52\% for SC-D, with
empirical SCER 5.38\%, 7.00\%, and 7.76\%, respectively.
Across the four CommonsenseQA models, SC-S adds 3.16--7.21 percentage points and
SC-D adds 6.43--10.28, with no loss of anchor-accepted answers in these runs.
On Qwen2.5-32B, SC-D adds 10.28 answers and 2.40 errors per 100 test-question
exposures. Their 23.37\% empirical error exceeds the target, while pooled SCER remains below it: reserve offsets incremental risk. Appendix~\ref{app:reserve-diagnostics} gives paired counts; Figure~\ref{fig:csqa-retention} covers all targets, and Table~\ref{tab:cross-dataset} shows ARC-Challenge gains.

The prespecified max-margin anchor supports reserve and window certification; it does not directly maximize final retention. KL certification uncertainty and worst-case expansion costs further restrict the window.
The prespecified coverage anchor raises SC-D retention to
99.97\% (empirical SCER 16.01\%) on the same configuration.
Appendix~\ref{app:reserve-diagnostics} separates these effects;
Figure~\ref{fig:sensitivity}(f) compares concentration bounds.
SC-D's wider windows in Figure~\ref{fig:certificate-geometry} reflect its specialized bound. TriviaQA Qwen2.5-32B has median margin 0.0165 and
direct radius 45, versus 0.0816 and 279 on CommonsenseQA. SC-S is
unavailable on all four TriviaQA models at $\alpha=0.20$. Error-ranking AUROCs
overlap (0.789--0.842 versus 0.799--0.846), so weak ranking alone
does not explain this outcome. Appendix~\ref{app:reserve-diagnostics}
separates margin and ranking diagnostics, and Appendix~\ref{app:followup-evaluation} gives question-disjoint evaluation.

\subsection{Exact Counterexample and Risk Transfer}
\label{sec:experiment-mechanism}
Figure~\ref{fig:lec-exact-counterexample} shows that complete-scan
LEC ($c=1$) violates the marginal target on the four-band i.i.d. law:
exact enumeration gives $\mathcal R_{25}(A_1)\approx0.002078>0$ and
SCER $\approx0.2080>0.20$. These calibration-averaged population quantities differ from the finite-pool QA estimates. The signed gap $0.024800$ exceeds reserve $0.022722$, so empirical feasibility alone does not ensure population risk control. Appendix~\ref{app:lec-counterexample}
gives the exact calculation for the published complete-scan rule.
On this small-$n$ law, SC-D retains the known safe anchor and SC-S abstains; neither expands. Expansion occurs on the QA benchmarks and the strong-core distribution in Appendix~\ref{app:followup-certification}.
Directional counting tightens the absolute bound by 27.3\% there, but neither bound certifies unrestricted LEC. Offline cost and routing analyses appear
in Appendices~\ref{app:followup-cost}--\ref{app:followup-routing}.

\section{Conclusion}
In this paper, we identify an exact i.i.d. counterexample that establishes a gap between empirical feasibility and marginal SCER control in LEC's adaptive threshold search.
Building on stability-based risk transfer, SC-LEC compensates for threshold-selection instability using the negative risk of a full-data reference.
Furthermore, SC-S develops bounds on this instability through local rank crossings, while SC-D directly bounds the risk of anchored expansion and permits wider windows under the specified shared-design conditions.
With valid population margins, both certificates ensure finite-sample marginal SCER control at positive retention.
Independent margin certification provides an upstream $1-\delta$ certificate of this calibration-averaged guarantee.
Empirical evaluations on three QA datasets demonstrate retention gains beyond the same safe anchor and show how anchor selection and small certified margins affect those gains.

\clearpage

\begingroup
\interlinepenalty=10000
\bibliography{custom}
\endgroup

\appendix


\clearpage

\section{Proofs}
\label{app:proofs}

Proves of Theorems~\ref{thm:decomposition}, \ref{thm:scer-control},
\ref{thm:explicit-scer}, \ref{thm:direct-anchor}, and
\ref{thm:certified-marginal-scer} are provided in
Sections~\ref{app:proof-decomposition}, \ref{app:proof-scer-control},
\ref{app:proof-explicit-scer}, \ref{app:proof-direct-anchor}, and
\ref{app:proof-certified-marginal-scer}, respectively.

We use the notation of Section~\ref{sec:method}. In particular,
$N=n+1$, $g_i(\lambda)=g_\alpha(W_i;\lambda)$, and
$T_m(\lambda)=\sum_{i=1}^m g_i(\lambda)$.
The full-data and leave-one-out algorithms share the buffer and, where
applicable, the anchor and radius.

\subsection{Linear Risk and Oracle Reserve}
\label{app:proof-linear-scer}
\label{app:proof-oracle-risk}

\paragraph{Linear Characterization.}

\begin{lemma}[Linear characterization]
\label{lem:linear-scer}
For positive marginal retention,
\begin{equation}
\operatorname{SCER}(\widehat\lambda)\le\alpha
\quad\Longleftrightarrow\quad \mathcal R_n(A)\le0.
\label{eq:appendix-linear-statement}
\end{equation}
\end{lemma}

\begin{proof}[Proof of Lemma~\ref{lem:linear-scer}]
When $\mathbb E[S_N(\widehat\lambda)]>0$, multiplying
Eq.~\eqref{eq:scer} by its positive denominator gives
\[
\begin{aligned}
&\operatorname{SCER}(\widehat\lambda)\le\alpha\\
&\quad\Longleftrightarrow
\mathbb E[Z_N(\widehat\lambda)]
\le\alpha\mathbb E[S_N(\widehat\lambda)]\\
&\quad\Longleftrightarrow
\mathbb E[g_N(\widehat\lambda)]\le0.
\end{aligned}
\]
The last expectation is $\mathcal R_n(A)$ by
Eq.~\eqref{eq:scer}.
\end{proof}

\paragraph{Oracle Safety Reserve.}

\begin{lemma}[Oracle safety reserve]
\label{lem:oracle-risk}
Under exchangeability and permutation invariance,
\begin{equation}
\begin{aligned}
\mathbb E[g_N(\widehat\lambda^\star)]
&=-\mathbb E[H_c(D_N)]\\
&\le-\frac{c}{N}\Pr(F_c).
\end{aligned}
\label{eq:oracle-risk}
\end{equation}
\end{lemma}

\begin{proof}[Proof of Lemma~\ref{lem:oracle-risk}]
The permutation invariance of $A_c$ and exchangeability of
$W_{1:N}$ imply
\[
\begin{aligned}
\mathbb E[g_N(\widehat\lambda^\star)]
&=\mathbb E\!\left[\frac1N\sum_{i=1}^N
g_i(\widehat\lambda^\star)\right]\\
&=-\mathbb E[H_c(D_N)].
\end{aligned}
\]
On $F_c$, feasibility gives $T_N(\widehat\lambda^\star)\le-c$. On $F_c^c$, the output is $\bot$ and this sum is zero. Thus
\begin{equation}
H_c(D_N)\ge\frac cN\mathbf 1\{F_c\}.
\label{eq:reserve-lower}
\end{equation}
Taking expectations proves Eq.~\eqref{eq:oracle-risk}.
The same argument applies to $A_{c,h}$ and its feasible set.
\end{proof}

\subsection{Margin--Stability Decomposition}
\label{app:proof-decomposition}
\label{app:shift-invariance}
\label{app:selective-loss-remark}

\begin{proof}[Proof of Theorem~\ref{thm:decomposition}]
Exchangeability and symmetry give
\[
\mathcal R_n(A_c)
=\mathbb E\!\left[\frac1N\sum_{i=1}^N
g_i(A_c(D_{-i}))\right].
\]
Adding and subtracting the full-data losses inside this expectation
yields
\[
\begin{aligned}
\mathcal R_n(A_c)
&=\mathbb E\!\left[\frac1N\sum_{i=1}^N
g_i(\widehat\lambda^\star)\right]+\Delta_c\\
&=-\mathbb E[H_c(D_N)]+\Delta_c.
\end{aligned}
\]
This proves Eq.~\eqref{eq:exact-decomposition}. Replacing $\Delta_c$ by an upper bound proves Eq.~\eqref{eq:decomposition-upper}.
The argument only uses symmetry and integrability, so it also applies
to the anchored rule.
\end{proof}

\paragraph{Shift Invariance.}

\begin{lemma}[Shift invariance]
\label{lem:shift-invariance}
For $\widetilde\ell_\alpha(W;\lambda)=g_\alpha(W;\lambda)+\alpha$,
\[
\begin{aligned}
&\widetilde\ell_\alpha(W;\lambda)
-\widetilde\ell_\alpha(W;\lambda')\\
&\qquad =g_\alpha(W;\lambda)-g_\alpha(W;\lambda').
\end{aligned}
\]
\end{lemma}
\begin{proof}
The common additive constant $\alpha$ cancels. In particular, this
identity also holds for $\bot$, whose shifted loss is $\alpha$.
\end{proof}

\paragraph{Remark: expected selective loss and marginal SCER.}
Let $S$ denote acceptance of a fresh test point by a calibration-fitted
rule and $Z=ES$. Our target is $\mathbb E[Z]/\mathbb E[S]$, with both
expectations over calibration and testing. For $r=\mathbb E[S]>0$,
\[
\begin{aligned}
\mathbb E[\widetilde\ell_\alpha]&=\alpha+\mathbb E[g_\alpha],\\
\operatorname{SCER}&=\alpha+\mathbb E[g_\alpha]/r.
\end{aligned}
\]
Thus $\mathbb E[\widetilde\ell_\alpha]\le\alpha+\beta$ alone leaves
SCER slack $\beta/r$. Recovering level $\alpha$ requires $\mathbb E[g_\alpha]\le0$.
In the preprint of \citet{angelopoulos2026conformal}, Eq.~(21) and Corollary~3 condition
on acceptance by the random calibrated threshold, not additionally on
the realized calibration sample: the marginalization agrees with ours.
Their Corollary~3 states selective accuracy at least $1-\alpha-\beta$.
SC-LEC targets the exact $\alpha$ endpoint by certifying sufficient
negative reserve and making the required quantities operational.
The conversion above does not by itself imply the cited selective-accuracy corollary.

\subsection{Directional Stability and SCER Control}
\label{app:proof-rank-stability}
\label{app:proof-finite-bound}
\label{app:proof-scer-control}

\paragraph{Directional Rank Stability.}

\begin{proof}[Proof of Proposition~\ref{prop:rank-stability}]
Let $V_i=r_N(U_i)$ be the full-data rank of the $i$th score and define
\[
\delta_i=S_i(\widehat\lambda_{-i})
-S_i(\widehat\lambda^\star).
\]
Since $g_i(\lambda)=(E_i-\alpha)S_i(\lambda)$,
\[
g_i(\widehat\lambda_{-i})-g_i(\widehat\lambda^\star)
=(E_i-\alpha)\delta_i.
\]
Set $\mathcal I_+=\{i:\delta_i=1\}$ and
$\mathcal I_-=\{i:\delta_i=-1\}$. If $i\in\mathcal I_+$, then
\[
r^\star<V_i\le r_i\le r^\star+K_c^+.
\]
If $i\in\mathcal I_-$, then
\[
r^\star-K_c^-\le r_i<V_i\le r^\star.
\]
The scores are distinct, so the $V_i$ form a permutation of $[N]$.
The two windows therefore give
\[
|\mathcal I_+|\le K_c^+,\qquad
|\mathcal I_-|\le K_c^-.
\]
Keeping the direction of each loss difference,
\[
\begin{aligned}
\sum_{i=1}^N(E_i-\alpha)\delta_i
&=\sum_{i\in\mathcal I_+}(E_i-\alpha)\\
&\quad +\sum_{i\in\mathcal I_-}(\alpha-E_i)\\
&\le(1-\alpha)K_c^++\alpha K_c^-\\
&\le K_c.
\end{aligned}
\]
Divide by $N$ and take expectations to obtain
Eq.~\eqref{eq:beta-rank-directional}.
\end{proof}

\paragraph{Generic Risk Bound and SCER Control.}

\begin{corollary}[Finite-sample linear-risk bound]
\label{cor:finite-bound}
Under the assumptions of Theorem~\ref{thm:decomposition} and
Proposition~\ref{prop:rank-stability},
\begin{equation}
\mathcal R_n(A_c)\le\beta_c-\frac{c}{N}\Pr(F_c).
\label{eq:main-risk-bound}
\end{equation}
\end{corollary}

\begin{proof}[Proof of Corollary~\ref{cor:finite-bound}]
Theorem~\ref{thm:decomposition} and
Proposition~\ref{prop:rank-stability} give
$\mathcal R_n(A_c)\le-\mathbb E[H_c(D_N)]+\beta_c$.
Eq.~\eqref{eq:reserve-lower} gives
$\mathbb E[H_c(D_N)]\ge(c/N)\Pr(F_c)$.
Combining these inequalities proves Eq.~\eqref{eq:main-risk-bound}.
\end{proof}

\paragraph{Stability-Compensated SCER Control.}

A simpler sufficient condition for Theorem~\ref{thm:scer-control} is
\begin{equation}
\mathbb E[K_c]\le c\Pr(F_c).
\label{eq:simple-main-condition}
\end{equation}
When $\Pr(F_c)=1$, this reduces to $\mathbb E[K_c]\le c$.

\begin{proof}[Proof of Theorem~\ref{thm:scer-control}]
Eq.~\eqref{eq:main-condition} is equivalent to
$N\beta_c\le c\Pr(F_c)$. Corollary~\ref{cor:finite-bound}
therefore gives $\mathcal R_n(A_c)\le0$.
Lemma~\ref{lem:linear-scer} gives the SCER guarantee when retention
is positive. Since
\[
(1-\alpha)\mathbb E[K_c^+]+\alpha\mathbb E[K_c^-]
\le\mathbb E[K_c],
\]
Eq.~\eqref{eq:simple-main-condition} is also sufficient.
\end{proof}

\subsection{KL Feasibility and Local-Crossing Analysis}
\label{app:proof-anchor-feasible}
\label{app:proof-local-stability}
\label{app:proof-explicit-scer}

\paragraph{KL Anchor-Feasibility Bound.}

The anchor-feasibility and anchored-search results below, including the supplementary rank bounds, use the i.i.d., prespecified-design, distinct-score, and safe-anchor assumptions of
Section~\ref{sec:explicit-instantiation}. The general decision-rule
extension states its own conditions.
Binary KL is
\[
\operatorname{kl}(x\Vert y)=x\log\frac{x}{y}
+(1-x)\log\frac{1-x}{1-y},
\]
with its continuous endpoint conventions.

\begin{lemma}[Anchor feasibility]
\label{lem:anchor-feasible}
Under Eq.~\eqref{eq:safe-core}, for $0\le b<m\eta$,
\begin{equation}
\begin{aligned}
&\Pr\{T_m(\lambda_0)>-b\}\le q_{m,b},\\
&q_{m,b}\le\exp\{-2m(\eta-b/m)^2\}.
\end{aligned}
\label{eq:anchor-feasibility-tail}
\end{equation}
\end{lemma}

\begin{proof}[Proof of Lemma~\ref{lem:anchor-feasible}]
Let $Y_i=g_i(\lambda_0)+\alpha\in[0,1]$ and set
$p=\alpha-\eta\in(0,1)$. The safe-core assumption gives
$\mathbb E[Y_i]\le p$. For any $t>0$, convexity on $[0,1]$ gives
$e^{tY_i}\le1-Y_i+Y_i e^t$, and hence
\[
\mathbb E[e^{tY_i}]
\le1-\mathbb E[Y_i]+\mathbb E[Y_i]e^t
\le1-p+pe^t.
\]
For $0\le b<m\eta$, set $a=\alpha-b/m$. Then $p<a<1$.
Independence and the exponential Markov inequality imply
\[
\begin{aligned}
&\Pr\{T_m(\lambda_0)>-b\}\\
&\quad=\Pr\left\{\sum_{i=1}^mY_i>ma\right\}\\
&\quad\le e^{-tma}(1-p+pe^t)^m.
\end{aligned}
\]
The right-hand side is minimized at
\[
t^*=\log\frac{a(1-p)}{p(1-a)}>0.
\]
Substitution gives $\exp\{-m\operatorname{kl}(a\Vert p)\}=q_{m,b}$.
We bound the moment-generating function by that of a Bernoulli variable. This does not require the $Y_i$ themselves to be Bernoulli.

For completeness, for fixed $p$ let $f(x)=\operatorname{kl}(x\Vert p)$.
Then $f(p)=f'(p)=0$ and
$f''(x)=1/[x(1-x)]\ge4$. Integrating twice from $p$ to $a$ gives
\[
\operatorname{kl}(a\Vert p)\ge2(a-p)^2.
\]
Since $a-p=\eta-b/m$, the second inequality in
Eq.~\eqref{eq:anchor-feasibility-tail} follows. For $b\ge m\eta$,
the convention $q_{m,b}=1$ supplies the trivial valid upper bound.
\end{proof}

\paragraph{Local-Crossing Stability.}

Recall $\widetilde q_{N,c}=q_{N,c+\alpha}$ and $p_{N,c}=1-q_{N,c}$.
\[
\Pr(F_{c,h})\ge p_{N,c},\qquad
\mathbb E[H_{c,h}]\ge(c/N)p_{N,c}.
\]

\begin{lemma}[Local-crossing stability]
\label{lem:direct-local-stability}
Suppose $c+\alpha<N\eta$ and $0\le h\le n$. With
$M_\alpha=\max\{\alpha,1-\alpha\}$, define
\[
\gamma_{N,c,h}
=M_\alpha\bigl[h+(N-h)\widetilde q_{N,c}\bigr].
\]
Then
\begin{equation}
\Delta_{c,h}\le\frac{\gamma_{N,c,h}}N.
\label{eq:direct-local-stability}
\end{equation}
\end{lemma}

\begin{proof}[Proof of Lemma~\ref{lem:direct-local-stability}]
Define
\[
\widetilde{\mathcal G}
=\{T_N(\lambda_0)\le-c-\alpha\}.
\]
For each $i$, $g_i(\lambda_0)\ge-\alpha$, so on this event,
\begin{equation}
\begin{aligned}
T_{-i}(\lambda_0)
&=T_N(\lambda_0)-g_i(\lambda_0)\\
&\le T_N(\lambda_0)+\alpha\le-c.
\end{aligned}
\label{eq:loo-anchor-sum}
\end{equation}
Thus the anchor is feasible on $D_N$ and all $D_{-i}$, and every
output is at least $\lambda_0$. Lemma~\ref{lem:anchor-feasible},
applied with $m=N$ and $b=c+\alpha<N\eta$, gives
\begin{equation}
\Pr(\widetilde{\mathcal G}^c)
\le q_{N,c+\alpha}=\widetilde q_{N,c}.
\label{eq:tight-good-prob}
\end{equation}

Let
\[
\delta_i=S_i(A_{c,h}(D_{-i}))-S_i(A_{c,h}(D_N)).
\]
On $\widetilde{\mathcal G}$, points with $U_i\le\lambda_0$ are
accepted by both rules, so $\delta_i=0$. Consider a point above
the anchor with full-data above-anchor rank $k>h$. All thresholds
in $\mathcal T_h(D_N)$ lie below $U_i$. Deleting that point leaves
the first $h$ above-anchor scores unchanged, so all thresholds in
$\mathcal T_h(D_{-i})$ also lie below $U_i$. This point is rejected
by both rules. Therefore only the first $h$ above-anchor points can
contribute, and
\[
\sum_{i=1}^N|\delta_i|\le h
\quad\text{on }\widetilde{\mathcal G}.
\]
Outside this event, the sum is bounded by $N$. Since
$|E_i-\alpha|\le M_\alpha$, the pointwise loss difference sum is
at most $M_\alpha h$ on $\widetilde{\mathcal G}$ and at most
$M_\alpha N$ on its complement. Hence
\[
\begin{aligned}
N\Delta_{c,h}
&\le M_\alpha h\Pr(\widetilde{\mathcal G})
 +M_\alpha N\Pr(\widetilde{\mathcal G}^c)\\
&=M_\alpha\bigl[h+(N-h)\Pr(\widetilde{\mathcal G}^c)\bigr]\\
&\le M_\alpha\bigl[h+(N-h)\widetilde q_{N,c}\bigr].
\end{aligned}
\]
This is Eq.~\eqref{eq:direct-local-stability}.
\end{proof}

\paragraph{Explicit Stability Certificate.}

\begin{proof}[Proof of Theorem~\ref{thm:explicit-scer}]
Since $\lambda_0\in\mathcal T_h(D_N)$,
\[
\begin{aligned}
\Pr(F_{c,h})&\ge\Pr\{T_N(\lambda_0)\le-c\}\\
&\ge1-q_{N,c}=p_{N,c}.
\end{aligned}
\]
The oracle-reserve argument in
Appendix~\ref{app:proof-oracle-risk} gives
$\mathbb E[H_{c,h}]\ge(c/N)p_{N,c}$.
Theorem~\ref{thm:decomposition} applies to $A_{c,h}$ because the
anchor is fixed and the candidate grid and maximization are symmetric.
Together with Lemma~\ref{lem:direct-local-stability}, this gives
\[
\begin{aligned}
\mathcal R_n(A_{c,h})
&=-\mathbb E[H_{c,h}]+\Delta_{c,h}\\
&\le\frac{\gamma_{N,c,h}-c\,p_{N,c}}N.
\end{aligned}
\]
Under Eq.~\eqref{eq:explicit-condition-tight}, this is nonpositive.
The SCER statement follows from Lemma~\ref{lem:linear-scer}.
\end{proof}

\subsection{Direct Anchored Risk Control}
\label{app:proof-direct-anchor}
\label{app:proof-retention}

\paragraph{Structural Refinement for Anchored Search.}

\begin{proof}[Proof of Theorem~\ref{thm:direct-anchor}]
Write $\widehat\lambda=A_{c,h}(D_{1:n})$ and
$Q=\{\widehat\lambda\ne\bot\}$. On $Q$, the selected threshold is
at least $\lambda_0$. With
\[
B=Q\cap\{\lambda_0<U_N\le\widehat\lambda\},
\]
the loss splits exactly as
\begin{equation}
g_N(\widehat\lambda)
=\mathbf 1\{Q\}g_N(\lambda_0)
 +(E_N-\alpha)\mathbf 1\{B\}.
\label{eq:app-anchor-expansion-split}
\end{equation}

\paragraph{Anchor contribution.}
The event $Q$ depends only on $D_{1:n}$. Since the test observation
is independent of that dataset and the anchor is fixed,
\[
\mathbb E[\mathbf 1\{Q\}g_N(\lambda_0)]
=\mu_0\Pr(Q).
\]
Feasibility of $\lambda_0$ implies $Q$, so
Lemma~\ref{lem:anchor-feasible} gives
$\Pr(Q)\ge1-q_{n,c}$. Using $\mu_0\le-\eta$,
\[
\mu_0\Pr(Q)\le-\eta(1-q_{n,c}).
\]

\paragraph{Expansion contribution.}
For any dataset $D$, let $t_h(D)=\max\mathcal T_h(D)$. In particular $t_0(D)=\lambda_0$, and if fewer than $h$ above-anchor
scores are present, use their maximum or $\lambda_0$ when none
are present. Since $\widehat\lambda\le t_h(D_{1:n})$ on $Q$,
\begin{equation}
B\subseteq\{\lambda_0<U_N\le t_h(D_{1:n})\}.
\label{eq:app-expansion-envelope}
\end{equation}
In the full dataset, consider the above-anchor rank $k$ of $U_i$.
If $k>h$, deleting $W_i$ leaves the first $h$ above-anchor scores
strictly below $U_i$, so $t_h(D_{-i})<U_i$. Hence
\[
\sum_{i=1}^N\mathbf 1\{\lambda_0<U_i\le t_h(D_{-i})\}\le h.
\]
This counting argument requires neither anchor feasibility nor
any condition on the errors. Exchangeability and the symmetry of
$t_h$ yield
\begin{equation}
\begin{aligned}
&\Pr\{\lambda_0<U_N\le t_h(D_{1:n})\}\\
&\quad=\mathbb E\!\left[
\frac1N\sum_{i=1}^N
\mathbf 1\{\lambda_0<U_i\le t_h(D_{-i})\}\right]\\
&\quad\le\frac hN.
\end{aligned}
\label{eq:app-expansion-probability}
\end{equation}
Because $E_N-\alpha\le1-\alpha$, Eqs.~\eqref{eq:app-expansion-envelope}
and \eqref{eq:app-expansion-probability} imply
\[
\mathbb E[(E_N-\alpha)\mathbf 1\{B\}]
\le\frac{(1-\alpha)h}{N}.
\]
Combine the two contributions in
Eq.~\eqref{eq:app-anchor-expansion-split} to prove
Eq.~\eqref{eq:direct-anchor-risk}. Condition
\eqref{eq:direct-anchor-condition} makes the bound nonpositive, so Lemma~\ref{lem:linear-scer} gives SCER control.
Here empirical feasibility is used only to obtain
$\Pr(Q)\ge1-q_{n,c}$. The expansion bound follows solely from the prespecified rank window.
\end{proof}

\paragraph{Nontrivial Retention and the Zero-Radius Case.}

\begin{corollary}[Nontrivial retention]
\label{cor:retention}
Under the assumptions of Theorem~\ref{thm:direct-anchor},
\begin{equation}
\mathbb E[S_N(A_{c,h}(D_{1:n}))]
\ge p_0(1-q_{n,c})>0.
\label{eq:retention-lower}
\end{equation}
This also applies to Theorem~\ref{thm:explicit-scer}, since its
condition $c+\alpha<N\eta$ and $\eta<\alpha$ imply $c<n\eta$.
\end{corollary}

\begin{proof}[Proof of Corollary~\ref{cor:retention}]
Let $Q_0=\{T_n(\lambda_0)\le-c\}$ denote calibration anchor
feasibility. On $Q_0$, the rule outputs a threshold at least
$\lambda_0$. Therefore
\[
S_N(A_{c,h}(D_{1:n}))
\ge\mathbf 1\{Q_0\}\mathbf 1\{U_N\le\lambda_0\}.
\]
The two indicators on the right are independent, so
\[
\begin{aligned}
\mathbb E[S_N(A_{c,h}(D_{1:n}))]
&\ge p_0\Pr(Q_0)\\
&\ge p_0(1-q_{n,c}).
\end{aligned}
\]
The safe-core assumption implies $p_0>0$. Since $c<n\eta$,
$\alpha-c/n>\alpha-\eta$, making the KL divergence strictly
positive and $q_{n,c}<1$.

For Theorem~\ref{thm:explicit-scer}, its assumption
$c+\alpha<N\eta$ implies
$c<n\eta+\eta-\alpha<n\eta$, so the same argument applies.
\end{proof}

For $h=0$, write $\widehat\lambda_0=A_{c,0}(D_{1:n})$ and
$Q=\{\widehat\lambda_0=\lambda_0\}$. Independence of the fixed
anchor losses and $Q$ gives
\[
\begin{aligned}
\mathbb E[Z_N(\widehat\lambda_0)]
&=\Pr(Q)\Pr(E=1,U\le\lambda_0),\\
\mathbb E[S_N(\widehat\lambda_0)]
&=\Pr(Q)p_0.
\end{aligned}
\]
If retention is positive, their ratio equals
$\Pr(E=1\mid U\le\lambda_0)$ exactly.

\subsection{Additional Stability Results}
\label{app:supplementary-rank}
\label{app:general-extension}

\paragraph{Supplementary Rank Bounds.}

Define $K_{c,h}$ by replacing $A_c$ with $A_{c,h}$ in the
definition of $K_c$. The following bounds control maximum rank displacement. Lemma~\ref{lem:direct-local-stability} instead counts observations whose acceptance changes, giving the sharper local stability certificate used in the main text.

\begin{lemma}[Union-bound rank stability]
\label{lem:local-rank-union}
Under the safe-core assumptions, suppose $c<n\eta$ and
$0\le h\le n$. Define
\begin{equation}
q^{\mathrm{ub}}_{n,c}
=\min\{1,q_{N,c}+Nq_{n,c}\}.
\label{eq:qnc}
\end{equation}
Then
\begin{equation}
\mathbb E[K_{c,h}]
\le(h+1)+(N-h-1)q^{\mathrm{ub}}_{n,c}.
\label{eq:kappa-bound}
\end{equation}
\end{lemma}
\begin{proof}
Let $\mathcal G$ be the event that the anchor is feasible on
$D_N$ and every $D_{-i}$. Applying Lemma~\ref{lem:anchor-feasible}
to each dataset and taking a union bound gives
$\Pr(\mathcal G^c)\le q^{\mathrm{ub}}_{n,c}$.
On $\mathcal G$, every output has full-data rank between the
anchor rank and that rank plus $h+1$: deleting one observation
can introduce at most one additional full-data score into the
local window. Hence $K_{c,h}\le h+1$ on $\mathcal G$.
Always $K_{c,h}\le N$. Taking expectations proves the claim.
\end{proof}

\begin{lemma}[Union-free rank stability]
\label{lem:local-rank-tight}
Under the assumptions of Lemma~\ref{lem:direct-local-stability},
\begin{equation}
\mathbb E[K_{c,h}]
\le(h+1)+(N-h-1)\widetilde q_{N,c}.
\label{eq:tight-rank-bound}
\end{equation}
\end{lemma}
\begin{proof}
Eq.~\eqref{eq:loo-anchor-sum} shows that
$\widetilde{\mathcal G}$ implies simultaneous anchor feasibility.
The same rank argument gives $K_{c,h}\le h+1$ on this event,
and $K_{c,h}\le N$ on its complement.
Eq.~\eqref{eq:tight-good-prob} bounds the complement probability
by $\widetilde q_{N,c}$, yielding the stated expectation bound.
\end{proof}

\paragraph{Extension to General Decision Rules.}

For a decision parameter $\theta$, let $S_i(\theta)$ and $Z_i(\theta)$
be binary indicators satisfying $Z_i(\theta)\le S_i(\theta)$.
Define $g_\alpha(W_i;\theta)=Z_i(\theta)-\alpha S_i(\theta)$.
For calibration and reference algorithms $A$ and $A^\star$, write
\begin{align*}
R&=\mathbb E[g_\alpha(W_N;A(D_{1:n}))],\\
R^\star&=\mathbb E[g_\alpha(W_N;A^\star(D_N))].
\end{align*}
If $R^\star\le-b$ and $R-R^\star\le\beta$, then
\[
R=R^\star+(R-R^\star)\le-b+\beta.
\]
Thus $\beta\le b$ gives SCER control when retention is positive.
For exchangeable data and symmetric algorithms, $R-R^\star$ also
has the leave-one-out representation
\[
\begin{aligned}
R-R^\star
&=\mathbb E\!\left[\frac1N\sum_{i=1}^N d_i\right],\\
d_i
&=g_\alpha(W_i;A(D_{-i}))\\
&\quad-g_\alpha(W_i;A^\star(D_N)).
\end{aligned}
\]
The decomposition is general, but a bound on this gap must be
established for the specific routing or multidimensional rule.

\subsection{Independent Margin Certification}
\label{app:proof-margin-certification}
\label{app:proof-certified-marginal-scer}

\paragraph{Simultaneous Anchor-Margin Certification.}

\begin{proof}[Proof of Eq.~\eqref{eq:simultaneous-anchor-certificate}]
Condition on $D_{\rm sel}$, so all $L$ candidate thresholds are fixed.
For each $j$, let $Y=g_\alpha(W;\lambda_j)+\alpha\in[0,1]$ have
mean $\theta_j=\alpha+\mu_j$. For $t<0$, the secant bound gives
\[
\mathbb E[e^{tY}]\le1-\theta_j+\theta_j e^t.
\]
Independence, exponential Markov, and optimization over $t<0$
therefore yield, for $0\le x<\theta_j<1$,
\begin{equation}
\Pr\{\overline Y_j\le x\mid D_{\rm sel}\}
\le e^{-m\operatorname{kl}(x\Vert\theta_j)}.
\label{eq:certification-lower-tail}
\end{equation}
The endpoint $x=0$ follows by a limit. If $\theta_j=0$ or $1$,
the bounded variable is constant, so the proposed upper bound covers
its mean almost surely.

Set $a=\log(L/\delta)/m$. For a fixed observation
$\overline Y_j=y<1$, the function
$u\mapsto\operatorname{kl}(y\Vert u)$ increases strictly on $(y,1)$,
because its derivative is $(u-y)/[u(1-u)]$.
Thus $\theta_j>u_j$ implies
\[
\overline Y_j<\theta_j,\qquad
\operatorname{kl}(\overline Y_j\Vert\theta_j)>a.
\]
For $0<\theta_j<1$, KL decreases continuously in its first argument
on $[0,\theta_j]$. If $a\ge\operatorname{kl}(0\Vert\theta_j)$,
the strict noncoverage event is empty. Otherwise, there is a unique
$x_a\in(0,\theta_j)$ with
$\operatorname{kl}(x_a\Vert\theta_j)=a$.
Noncoverage implies $\overline Y_j<x_a$. Applying
Eq.~\eqref{eq:certification-lower-tail} at $x_a$ gives
\[
\Pr\{\theta_j>u_j\mid D_{\rm sel}\}\le e^{-ma}=\delta/L.
\]
A union bound over $j$ gives simultaneous coverage with conditional
probability at least $1-\delta$. No independence across candidates
is needed. Averaging over $D_{\rm sel}$ proves
Eq.~\eqref{eq:simultaneous-anchor-certificate}.

On this simultaneous event, any candidate selected using the
certification data satisfies $\mu_{j^\star}\le u_{j^\star}-\alpha$.
For $u_{j^\star}<\alpha$, this is $\mu_0\le-\widehat\eta$.
For finite $m$ and $0<\delta<1$, the KL upper endpoint is positive,
including when $\overline Y_j=0$. Thus a passing margin is strictly less than $\alpha$, as required by the anchored theorems.
\end{proof}

\paragraph{Certified Marginal SCER and Unconditional Risk.}

\begin{proof}[Proof of Theorem~\ref{thm:certified-marginal-scer}]
Let $G$ denote simultaneous coverage in
Eq.~\eqref{eq:simultaneous-anchor-certificate}. Then $\Pr(G^c)\le\delta$.
Fix an upstream realization $H\in G$. If the procedure does not
certify a design, it abstains identically and $\mathcal R(H)=r(H)=0$.
Otherwise, its anchor has population margin at least
$\widehat\eta>0$, and its parameters satisfy either
Eq.~\eqref{eq:explicit-condition-tight} or
Eq.~\eqref{eq:direct-anchor-condition}, evaluated using that margin.
Since $D_{\rm cal}$ and the fresh test point remain i.i.d. and
independent of $H$, conditioning on $H$ makes the design deterministic
and preserves the assumptions of the corresponding theorem.
Therefore $\mathcal R(H)\le0$.

For a passing design, let $p_0(H)$ denote the population retention
of its selected anchor.
The pointwise bound
$g_\alpha(W;\lambda_0)\ge-\alpha\mathbf 1\{U\le\lambda_0\}$ and
$\mu_0\le-\widehat\eta$ give
\[
p_0(H)\ge\widehat\eta/\alpha.
\]
Corollary~\ref{cor:retention}, applied conditionally on $H$, gives
$r(H)\ge p_0(H)(1-q_{n,c})$.
The direct rule has $c<n\widehat\eta$, and the stability rule has
$c+\alpha<N\widehat\eta$, which also implies $c<n\widehat\eta$.
Thus $q_{n,c}<1$ and the retention bound is strictly positive.
Lemma~\ref{lem:linear-scer} gives $\operatorname{SCER}_H\le\alpha$.
Every failure of these conclusions for a passing design is contained
in $G^c$, proving the theorem.
\end{proof}

To obtain the unconditional bound, use the tower property and
$\mathcal R(H)\le1-\alpha$ for every $H$:
\begin{equation}
\begin{aligned}
\mathbb E[g_\alpha(W;\widehat\lambda)]
&=\mathbb E[\mathcal R(H)]\\
&\le(1-\alpha)\Pr(G^c)\\
&\le(1-\alpha)\delta.
\end{aligned}
\label{eq:app-certification-unconditional-risk}
\end{equation}
For positive overall retention, the identity
\[
\begin{aligned}
\operatorname{SCER}_{\rm all}
&=\alpha+
\frac{\mathbb E[g_\alpha(W;\widehat\lambda)]}
{\mathbb E[S(W;\widehat\lambda)]}
\end{aligned}
\]
gives the retention-normalized bound
\begin{equation}
\operatorname{SCER}_{\rm all}\le\alpha+
\frac{(1-\alpha)\delta}{\mathbb E[S(W;\widehat\lambda)]}.
\label{eq:certification-unconditional-scer}
\end{equation}
The confidence statement concerns the upstream data $H$. Conditional on these data, the risk bound still averages over final calibration $D_{\rm cal}$ and a fresh test observation. In particular, the
$q_{n,c}$ and $q_{N,c}$ terms in the anchored theorems are already
integrated into their marginal-risk bounds and are not additional
certification failure probabilities to be added to $\delta$.

\subsection{Failure-Abstention Counterexample}
\label{app:certification-implementation}

Consider $\alpha=0.2$, a single candidate $\lambda_0=1$, and
$U\sim\operatorname{Unif}(0,1)$ independent of
$E\sim\operatorname{Bernoulli}(0.4)$. Thus every point belongs
to the anchor, whose SCER is $0.4$ and whose linear risk is $0.2$.
With $m=20$ and $\delta=0.05$, an all-correct certification sample
has $\overline Y=0$ and
\[
u=1-0.05^{1/20}\approx0.1391<\alpha.
\]
It falsely certifies a positive margin, with positive probability
$0.6^{20}$. On this outcome, the direct default uses $c=0$.
The candidate window contains only $\lambda_0$, because there are
no scores above it. An all-correct final calibration sample makes
the anchor feasible, so the entire procedure has positive retention.
Whenever it accepts a test point, its decision depends only on
upstream and calibration data, independently of the test error.
Hence its unconditional SCER is exactly $0.4$, even though the
procedure abstains whenever certification does not pass. It exceeds
both $\alpha$ and $\alpha+\delta$. This example is compatible with
Eq.~\eqref{eq:app-certification-unconditional-risk}: overall retention
is small, and the error budget must be scaled by retention when
converted to SCER.

\section{Experimental Details and Additional Results}
\label{app:experiments}
\raggedbottom
\SCappendixfloats
\setcounter{figure}{0}
\renewcommand{\thefigure}{A\arabic{figure}}

\subsection{Datasets, Models, and Splits}
\label{app:server-protocol}
For paired comparisons, all four predictors use a common CommonsenseQA
pool fixed before inference: 10,956 records yield 10,955 questions after
normalized-question deduplication.
ARC-Challenge pools labeled train/validation/test splits
(1,119/299/1,172 rows). The 2,590 rows become 2,579 questions.
Closed-book TriviaQA uses the retained validation pool without
retrieved evidence: 17,944 retrieval rows become 9,961 question
records. Aliases are merged only for the same question ID and
normalized question. One reused ID with two different questions
remains two records. Normalized-question deduplication precedes
splitting and is independent of prediction errors. These are fixed
labeled evaluation pools, not official test submissions.

In each of 100 repetitions, we assign 10\%, 35\%, and 25\% of the benchmark pool to selection, certification, and final calibration, rounding each count down. The remaining questions form the test set. Exact counts are
1,095/3,834/2,738/3,288 for CommonsenseQA,
257/902/644/776 for ARC, and 996/3,486/2,490/2,989 for TriviaQA.
Primary selection uses scores only. Anchor uses $m$ certification
labels, LEC uses $n$ calibration labels, and SC-S/SC-D use $m+n$
labels in independent roles.
Models and methods share question-role assignments within each
fixed allocation. Selection determines the candidate grid and any
fusion standardization. Neither final calibration nor test data sets
the certified anchor, $\widehat\eta,c$, or $h$.
The $\delta=0.05$ budget applies to each model/score/target certificate,
not jointly to the entire collection of reported configurations.

All predictions use Qwen2.5-7B/32B-Instruct and Qwen3-8B/32B.
The prompts and generation settings are specified below.
Reported predictions follow this common protocol. Exploratory configurations are excluded, and failed configurations retain the prescribed abstention outcome.
\label{app:server-cached}

\subsection{Prediction and Uncertainty Extraction}
\label{app:server-scores}
Inference uses BF16 weights, batch size one, FP32 log-softmax, and
FP64 scalar summaries on A800 80GB GPUs. All prepared inputs fit within the 4,096-token limit, so no input truncation or exclusion was needed. Each model receives one user question, no demonstrations,
and its official chat template. Qwen3 thinking is disabled.
The exact user prompts appear in Appendix~\ref{app:prompts}. Choices are mapped in original order to
A/B/\ldots, including ARC's originally numeric labels. Each label's
fixed completion-token log-likelihood is summed across its tokens,
then normalized by softmax over labels. We predict the label with the highest normalized probability.
Primary uncertainty is $-\sum_k p_k\log p_k/\log K$, with $K$
choices. Logit fusion averages selection-standardized entropy,
predicted-label surprisal, and inverse top-two probability margin.

TriviaQA generates at most 36 greedy answer tokens. Terminal EOS
is excluded from likelihood summaries. Mean answer-token NLL and
predictive token entropy are obtained by teacher-forcing the generated
answer at the correct causal positions. Token fusion averages their
selection-standardized values. Correctness is normalized exact match
against all original and normalized aliases, including article and
punctuation normalization. Empty answers are counted as errors and assigned a fixed high uncertainty score. No example is excluded on the basis of correctness.

Table~\ref{tab:server-audit} summarizes prediction accuracy and score ties before question deduplication. Empirically distinct scores do not imply a
population atomlessness assumption: saturated probabilities can have
tiny entropy differences that disappear after rounding. Independent
uniform keys resolve actual ties without numerical jitter. Formally,
thresholds act on the lexicographic order of $(U,V)$, where $V$ is
an independent continuous uniform key attached to each observation.
The key stays fixed across roles and full/leave-one-out comparisons.
This gives an almost surely strict order without assuming that the
original score $U$ is atomless. The rank and crossing arguments use this augmented order.
Across the tested 12-decimal quantization configurations, changes in
SC-S/SC-D retention or empirical SCER are at most 0.008 percentage
points. Quantization occurs before every role, with fresh certification
and radius design.
\label{app:followup-precision}

\begin{table*}[!tbp]
\centering
\begingroup
\footnotesize
\begin{tabular}{llrrrrrr}
\SCtabletop
\rowcolor{SCHead}
Dataset & Model & Rows & Acc. (\%) & Unique (\%) & Max tie & 12dp (\%) & 12dp tie \\
\SCtablemid
\rowcolor{SCGroup}
 & Qwen2.5-7B & 10956 & 80.39 & 99.93 & 2 & 85.04 & 14 \\
\rowcolor{SCGroup}
 & Qwen2.5-32B & 10956 & 83.99 & 100.00 & 1 & 56.95 & 1507 \\
\rowcolor{SCGroup}
 & Qwen3-8B & 10956 & 78.07 & 99.94 & 2 & 72.98 & 411 \\
\rowcolor{SCGroup}
\multirow{-4}{*}{CommonsenseQA} & Qwen3-32B & 10956 & 81.14 & 100.00 & 1 & 99.80 & 2 \\
 & Qwen2.5-7B & 2590 & 89.61 & 99.42 & 2 & 86.87 & 6 \\
 & Qwen2.5-32B & 2590 & 95.29 & 99.34 & 2 & 35.48 & 815 \\
 & Qwen3-8B & 2590 & 90.77 & 99.46 & 2 & 60.54 & 222 \\
\multirow{-4}{*}{ARC-Challenge} & Qwen3-32B & 2590 & 95.25 & 99.73 & 2 & 97.14 & 3 \\
\rowcolor{SCGroup}
 & Qwen2.5-7B & 9961 & 45.28 & 99.57 & 32 & 99.53 & 32 \\
\rowcolor{SCGroup}
 & Qwen2.5-32B & 9961 & 61.02 & 97.09 & 74 & 96.20 & 74 \\
\rowcolor{SCGroup}
 & Qwen3-8B & 9961 & 46.26 & 98.69 & 37 & 98.15 & 37 \\
\rowcolor{SCGroup}
\multirow{-4}{*}{TriviaQA} & Qwen3-32B & 9961 & 50.02 & 99.98 & 2 & 99.98 & 2 \\
\SCtablebottom
\end{tabular}
\endgroup

\caption{Prediction accuracy and uncertainty-score ties before question deduplication. Unique reports the percentage of distinct scores, and Max tie is the largest tie group. The 12dp columns repeat these statistics after rounding scores to 12 decimal places.}
\label{tab:server-audit}
\end{table*}

\subsection{Full Risk-Target Results}
\label{app:server-results}
\begin{figure*}[!tbp]
\centering\includegraphics[width=\textwidth]{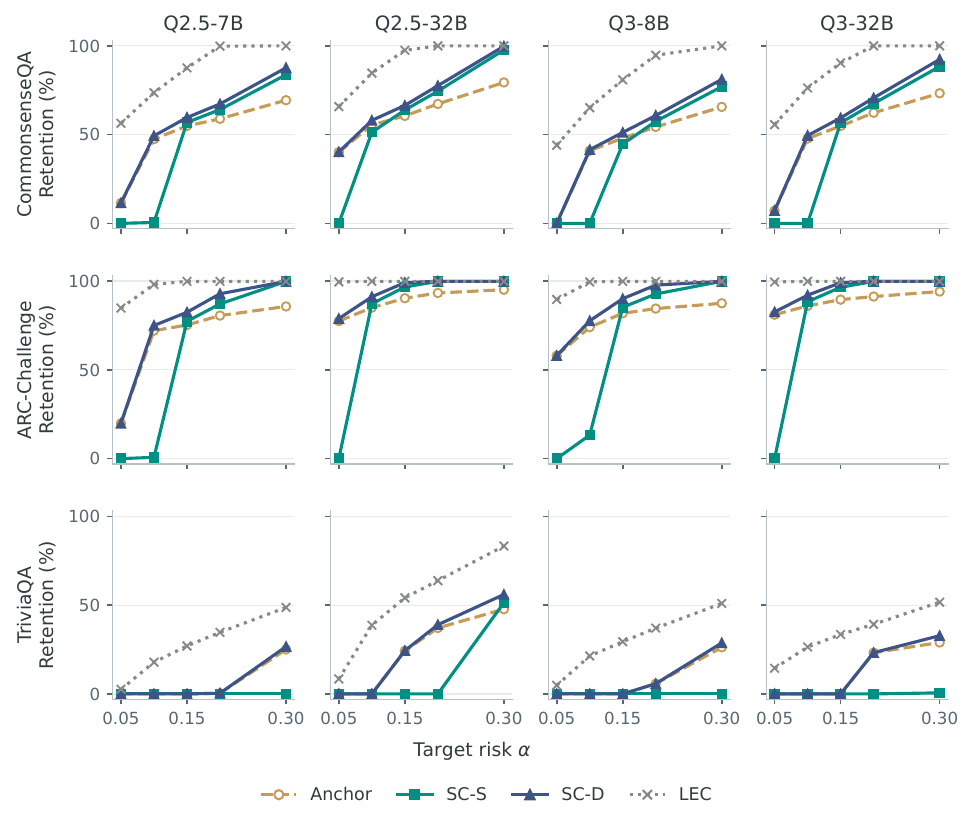}
\caption{Retention across all five target risks and twelve dataset--model pairs under the margin/KL design with primary uncertainty scores. Results average all 100 repetitions, including failed and unavailable designs. Zero retention corresponds to undefined SCER.}
\label{fig:full-kl-sweep}
\end{figure*}

\begin{figure*}[!tbp]
\centering\includegraphics[width=\textwidth]{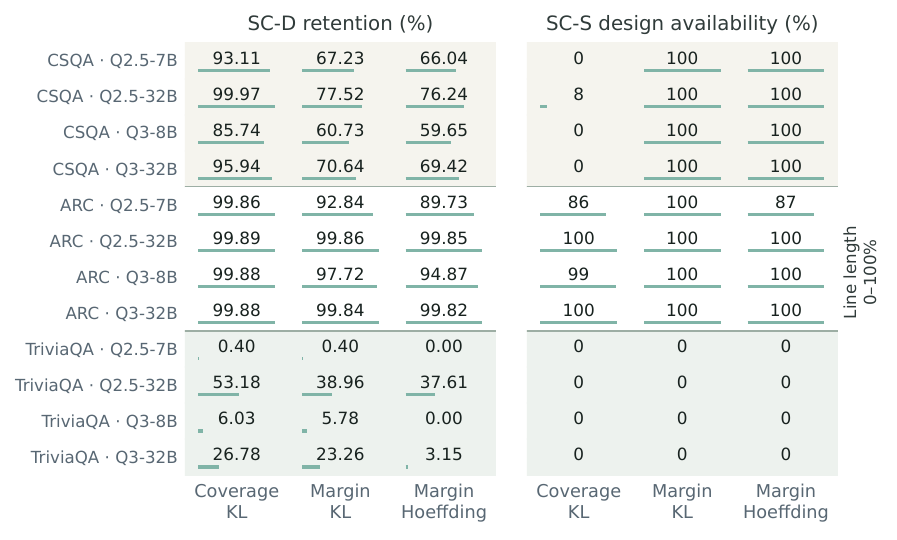}
\caption{Effects of anchor selection and concentration bounds at $\alpha=0.20$. The left panel reports SC-D retention, and the right panel reports the percentage of repetitions with an available SC-S design. Values are aggregated over all 100 repetitions. Numbers and horizontal bars use the same 0--100\% scale. The Hoeffding variant replaces both margin-certification and anchor-feasibility bounds.}
\label{fig:anchor-concentration}
\end{figure*}

Figure~\ref{fig:full-kl-sweep} reports all five target risks for the primary scores under the prespecified margin/KL design.
LEC's empirical risk is not interpreted as a certificate.
The coverage rule selects the largest passing anchor, whereas the margin rule selects the candidate with greatest certified reserve.
Figure~\ref{fig:anchor-concentration} compares these rules and a
combined Hoeffding~\citep{hoeffding1963probability} substitution for both certification and final
calibration tails. The latter is not a tail-only ablation.
The right panel reports how often an SC-S design is available. These comparisons use the primary scores; fusion-score analyses use the same protocol. All primary-score failures remain included.

\subsection{Certification and Sample-Size Sensitivity}
\label{app:followup-budget}
\begin{figure*}[!tbp]
\centering\includegraphics[width=\textwidth]{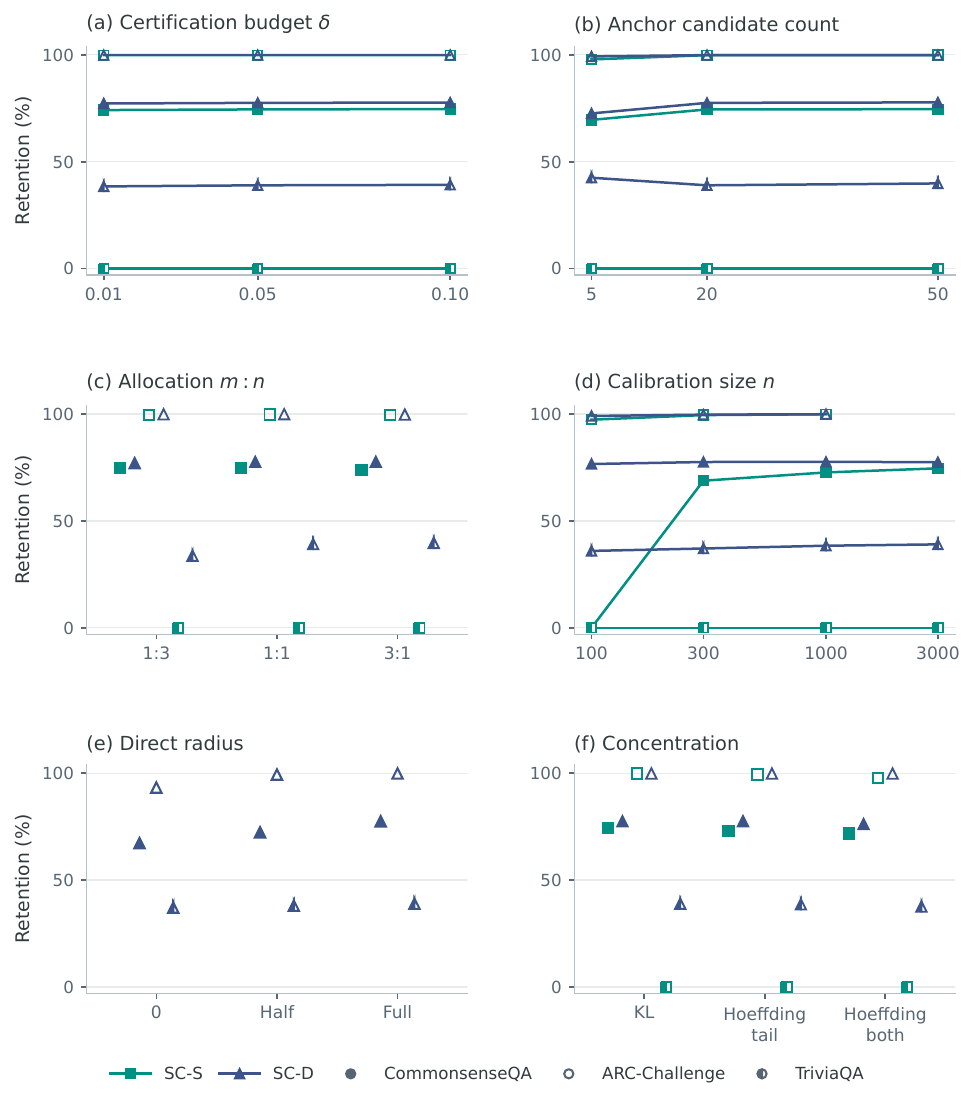}
\caption{Sensitivity of certified expansion for Qwen2.5-32B at $\alpha=0.20$. Retention includes all failed and unavailable designs, with undefined SCER at zero retention. Panels vary the certification error budget, candidate count, sample allocation, calibration size, SC-D radius, and concentration bounds. Panel (e) changes SC-D only. Panel (f) compares KL bounds with Hoeffding replacements for the tail alone or for both certification and tail bounds. Marker fill identifies the dataset, while color and shape identify the method. Small horizontal offsets separate coincident points. Allocation changes may also change the test set.}
\label{fig:sensitivity}
\end{figure*}

Figure~\ref{fig:sensitivity} compares Qwen2.5-32B across all three tasks
at $\alpha=0.20$, using primary scores and 100 repetitions.
These prespecified sensitivity analyses examine how calibration choices affect finite-sample conservatism.

Panels (a)--(b) vary $\delta$ and candidate count, which affect both
the simultaneous penalty and the selected anchor. More candidates
need not improve retention. An additional selection-single variant
uses selection labels to choose the largest empirically safe candidate,
then certifies it independently. It changes selection policy and
multiplicity, and costs extra labels beyond the default
certification-plus-calibration budget. Because the $L=1$ variant changes both anchor selection and the label budget, it does not isolate candidate multiplicity alone.

Panel (c) holds selection/test fractions at 0.10/0.30 and total
certification plus calibration at 0.60, changing only $m:n$ up to
rounding. Panel (d) fixes certification size while changing final
calibration size. Because the remaining test cohort changes size, these allocations do not use an identical evaluation set. ARC has
insufficient questions for $n=3000$, which is omitted.
For CommonsenseQA Qwen2.5-32B, SC-S changes from unavailable at
$n=100$ to 74.58\% retention at $n=3000$.

Panel (e) changes only SC-D's window. SC-S serves as a reference. An additional positive-buffer variant sets $c_D=0.25n\widehat\eta$ and recomputes a valid radius.
Panel (f) separates Hoeffding tails with KL certification from the
combined substitution, exposing two sources of conservatism.

The main-paper geometry in Figure~\ref{fig:certificate-geometry}
summarizes certified margin versus normalized expansion radius. The sensitivity analyses here explain how sample allocation and concentration
choice affect this geometry. At $\alpha=0.20$, CommonsenseQA Qwen2.5-32B
has mean SC-S/SC-D radii 195.55/279.12 ranks. TriviaQA's smaller margins
make SC-S unavailable. Distinct scores alone do not ensure useful certification. The KL bounds support larger margins/windows than the
combined Hoeffding replacement in these settings. At a fixed label
budget, changing TriviaQA Qwen2.5-32B's $m:n$ from $1:3$ to $3:1$
raises anchor certification from 87\% to 100\% and SC-D retention from
33.74\% to 39.72\%. These variants diagnose conservatism and do not
replace the main protocol after testing.

\subsection{Expansion and Independent Evaluation}
\label{app:followup-evaluation}
\begin{figure*}[!tbp]
\centering\includegraphics[width=\textwidth]{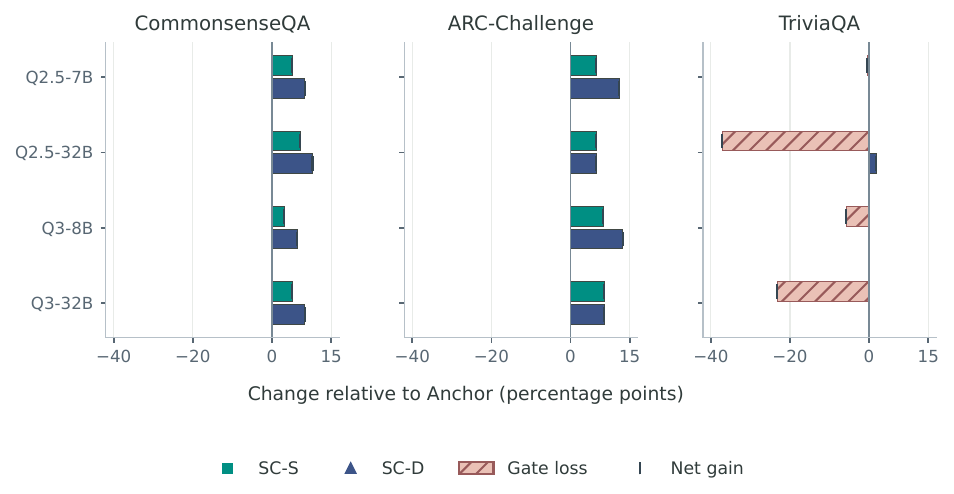}
\caption{Retention changes relative to the same anchor at $\alpha=0.20$ under the margin/KL design. Positive bars show additional accepted answers, hatched negative bars show lost anchor acceptance, and black ticks show the net change. Results include all 100 repetitions and are reported in percentage points.}
\label{fig:expansion-accounting}
\end{figure*}

Expansion beyond the same anchor and losses through the calibration
gate jointly determine net utility (Figure~\ref{fig:expansion-accounting}).
On TriviaQA, SC-S is unavailable and falls back to abstention. It therefore rejects answers that the anchor would accept, producing a negative net change in retention. This comparison does not
establish superiority over other calibration or anchor-selection rules.

\begin{figure*}[!tbp]
\centering\includegraphics[width=\textwidth]{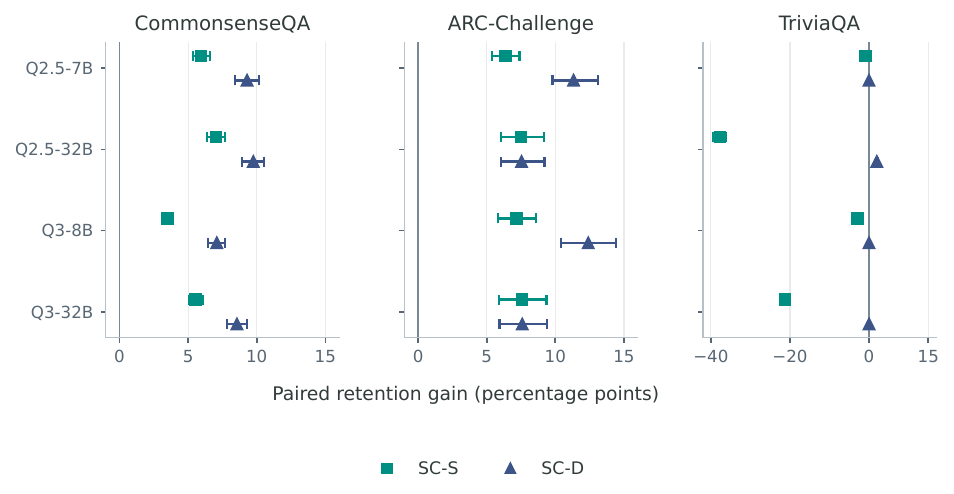}
\caption{Separate question-disjoint evaluation of paired retention gains over the same margin/KL anchor at $\alpha=0.20$. Pointwise 95\% intervals jointly bootstrap test questions while holding the 100 fitted policies fixed. They quantify question-level variability conditional on these policies, excluding refitting and certification uncertainty. Unavailable designs are included, giving negative TriviaQA SC-S gains. Dataset panels use their labeled horizontal ranges.}
\label{fig:holdout-forest}
\end{figure*}

\begin{figure*}[!tbp]
\centering\includegraphics[width=\textwidth]{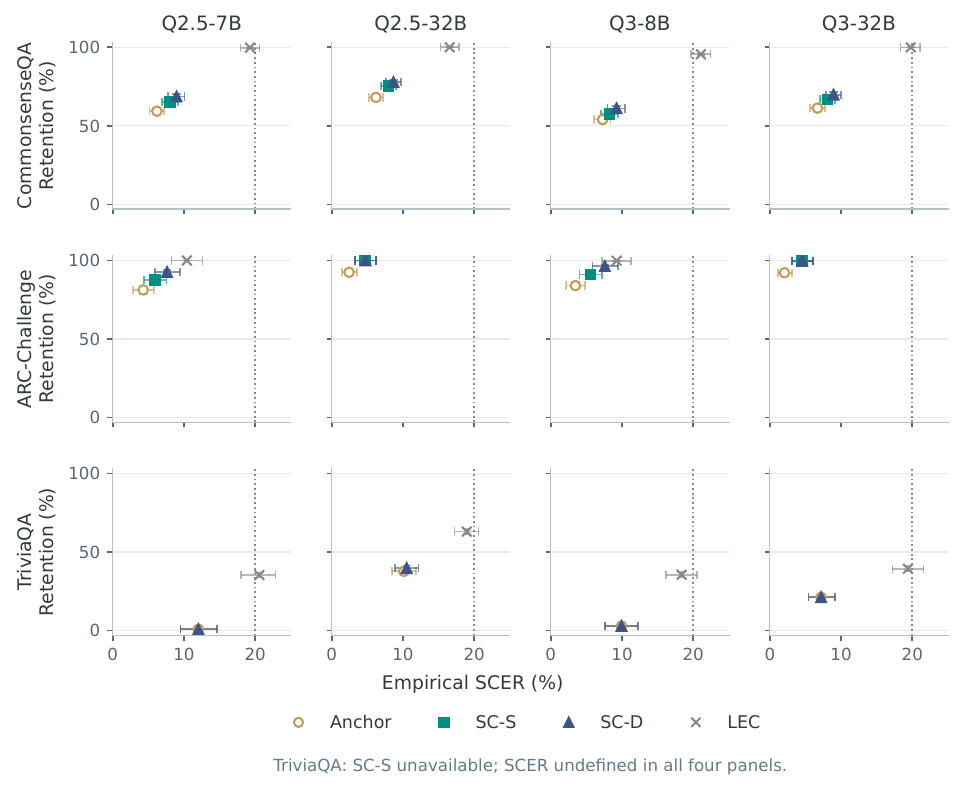}
\caption{Separate question-disjoint evaluation at $\alpha=0.20$, margin/KL, over all 100 fixed policies. Colors and shapes identify methods, and each panel represents a dataset/model pair. Horizontal and vertical bars are pointwise 95\% intervals from jointly bootstrapping test questions, conditional on these policies and excluding refitting and certification uncertainty. The vertical line marks the target. TriviaQA SC-S has zero retention and undefined SCER, shown by an annotation. LEC is an empirical reference.}
\label{fig:holdout-frontier}
\end{figure*}

The evaluation questions are disjoint from those used to fit these policies. However, the benchmark had already been used in exploratory analyses. The intervals therefore describe evaluation variability conditional on the fitted policies, rather than performance on a wholly untouched benchmark. Test size and remaining
role counts match the original allocation, and all 100 policies are fixed
before evaluation. We average acceptance across policies per question and
jointly bootstrap test questions for all methods 2,000 times. Pointwise
95\% percentile intervals describe question variability conditional on
these policies, not refitting variability or PAC certificate failure.
Figure~\ref{fig:holdout-forest} reports paired gains, and
Figure~\ref{fig:holdout-frontier} jointly shows retention and empirical SCER.
For CommonsenseQA Qwen2.5-32B, paired SC-S/SC-D gains are
7.05 [6.39, 7.70] and 9.75 [8.90, 10.55] percentage points.

\subsection{Accepted Errors and Certified Reserve}
\label{app:reserve-diagnostics}
Table~\ref{tab:reserve-diagnostics} examines every primary-score
configuration at $\alpha=0.20$ using the same 100 paired repetitions
as Table~\ref{tab:cross-dataset}. The fixed thresholds are evaluated
on their paired test questions. SC-D's acceptance contains Anchor's
on every evaluated question in these runs. Thus
$\Delta S=S_{\rm D}-S_{\rm A}\in\{0,1\}$ and
$\Delta Z=E\Delta S$, with
$\Delta g=\Delta Z-\alpha\Delta S$. We use these paired differences to describe the empirical effect of expansion.

The extra answers are more error-prone than the anchor answers.
On CommonsenseQA Qwen2.5-32B, the 100 repetitions contain 33,789
additional accepted answers and 7,896 additional errors over 328,800
test-question exposures. Their incremental error rate is 23.37\%,
while overall empirical SCER rises from 5.38\% to 7.76\%.
The mean linear loss increases by 0.00346 but remains negative.
Because questions recur across repetitions, exposures outnumber distinct questions and are not independent new observations. For SC-S, we distinguish added answers from lost anchor acceptance. At $\alpha=0.20$, SC-S is unavailable on all four TriviaQA models, so its negative net gain reflects lost anchor answers rather than nested expansion. Table~\ref{tab:reserve-diagnostics} reports the paired expansion counts.

The anchor objective, margin certification, and expansion bound each contribute to conservatism. The main
policy maximizes certified margin, whereas the coverage ablation takes
the largest passing anchor. Maximizing reserve need not maximize retention. Certification then controls joint linear risk, which
combines retention and conditional error. For CommonsenseQA
Qwen2.5-32B, the anchor's test linear loss is $-0.09833$. Its mean empirical negative loss on certification data is
0.09932. The mean KL allowance of 0.01765 reduces it to a mean certified margin of 0.08167. Because certification selects the anchor,
its empirical loss is a selection-dependent diagnostic, not an
independent estimate of the true margin. The design uses only the
certified lower bound, and the worst-case expansion allowance and,
for SC-S, simultaneous deleted-sample feasibility further limit the
window. Figure~\ref{fig:sensitivity}(f) contrasts KL with Hoeffding
tails alone and with both bounds replaced. Certification uncertainty
is one constraint, rather than an explanation of the entire gap
between observed SCER and the target.

TriviaQA illustrates a different bottleneck. Anchor certification
passes in 2, 100, 28, and 97 of 100 repetitions for Qwen2.5-7B/32B
and Qwen3-8B/32B, respectively. Conditional median margins are
0.00056, 0.01649, 0.00114, and 0.00411, with median direct radii
0, 45, 0, and 1. For Qwen2.5-32B, mean negative certification loss
0.03898 becomes a certified margin of 0.01667 after a KL allowance
of 0.02231. No SC-S design passes. Error-ranking AUROC uses the
complete deduplicated score/error pool, with larger uncertainty
predicting error and average ranks for ties. TriviaQA's AUROCs
overlap those of CommonsenseQA. The overlapping AUROC ranges show that global discrimination alone does
not measure the low-error prefix reserve. Here, the small certified
margins limit the permitted expansion.

\begin{table*}[!tbp]
\centering
\begingroup
\footnotesize
\begin{tabular}{llrrrrrrrrr}
\SCtabletop
\rowcolor{SCHead}
\shortstack{Dataset} & Model & AUROC & $100\widehat\eta$ & Pass & SC-S & $h_{\rm D}$ & $\Delta S$ & $\Delta Z$ & $q_{\rm new}$ & $100\Delta g$\\
\SCtablemid
\rowcolor{SCGroup}
 & Qwen2.5-7B & 0.827 & 6.72 & 100 & 100 & 230 & 8.35 & 2.11 & 25.27 & 0.44 \\
\rowcolor{SCGroup}
 & Qwen2.5-32B & 0.846 & 8.16 & 100 & 100 & 279 & 10.28 & 2.40 & 23.37 & 0.35 \\
\rowcolor{SCGroup}
 & Qwen3-8B & 0.799 & 5.19 & 100 & 100 & 177 & 6.43 & 1.46 & 22.66 & 0.17 \\
\rowcolor{SCGroup}
\multirow{-4}{*}{CSQA} & Qwen3-32B & 0.829 & 6.73 & 100 & 100 & 230 & 8.37 & 1.99 & 23.83 & 0.32 \\
\cmidrule(lr){1-11}
 & Qwen2.5-7B & 0.876 & 9.89 & 100 & 100 & 79 & 12.28 & 3.94 & 32.10 & 1.49 \\
 & Qwen2.5-32B & 0.921 & 13.96 & 100 & 100 & 112 & 6.56 & 2.29 & 34.88 & 0.98 \\
 & Qwen3-8B & 0.892 & 10.95 & 100 & 100 & 88 & 13.22 & 5.00 & 37.84 & 2.36 \\
\multirow{-4}{*}{ARC} & Qwen3-32B & 0.922 & 14.13 & 100 & 100 & 113 & 8.57 & 2.89 & 33.75 & 1.18 \\
\cmidrule(lr){1-11}
\rowcolor{SCGroup}
 & Qwen2.5-7B & 0.823 & 0.06 & 2 & 0 & 0 & 0.00 & 0.00 & -- & 0.00 \\
\rowcolor{SCGroup}
 & Qwen2.5-32B & 0.842 & 1.65 & 100 & 0 & 45 & 1.81 & 0.37 & 20.34 & 0.01 \\
\rowcolor{SCGroup}
 & Qwen3-8B & 0.830 & 0.11 & 28 & 0 & 0 & 0.00 & 0.00 & -- & 0.00 \\
\rowcolor{SCGroup}
\multirow{-4}{*}{TriviaQA} & Qwen3-32B & 0.789 & 0.41 & 97 & 0 & 1 & 0.07 & 0.02 & 23.27 & 0.00 \\
\SCtablebottom
\end{tabular}
\endgroup

\caption{Anchor margins, certificate availability, and incremental errors at $\alpha=0.20$, margin/KL.
AUROC uses the complete deduplicated benchmark pool (higher score
predicts error). $\widehat\eta$ is the median certified margin
conditional on passing. Pass and SC-S give anchor passes and
available SC-S designs out of 100. $h_{\rm D}$ is the median radius
conditional on an available direct design. $\Delta S$ and $\Delta Z$
are additional SC-D answers and errors per 100 test exposures. $q_{\rm new}=100\sum\Delta Z/\sum\Delta S$ is undefined when no
answers are added. The final column is the change in mean linear
loss per 100 exposures. All failed designs are included. SC-D loses no anchor-accepted answers in these runs.}
\label{tab:reserve-diagnostics}
\end{table*}

\subsection{Known-Distribution Validation}
\label{app:exp-ttt}
\label{app:followup-certification}
\begin{figure*}[!tbp]
\centering
\includegraphics[width=\textwidth]{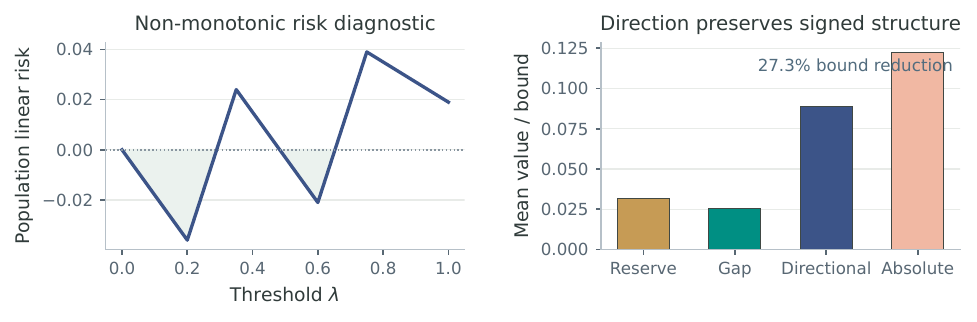}
\caption{Reserve and stability bounds for unrestricted LEC with $c=1$ under the distribution defined in Appendix~\ref{app:followup-certification}. Left: the integrated population linear-risk curve. Right: mean reserve, signed gap, directional bound, and absolute bound across 500 full-data/leave-one-out evaluations of the same rule. The directional bound is 27.3\% smaller than the absolute bound, but neither certifies the unrestricted rule.}
\label{fig:ttt-mechanism}
\end{figure*}

\paragraph{Distribution for stability analysis.}
Figure~\ref{fig:ttt-mechanism} shows a separate stability diagnostic.
We draw $U\sim\operatorname{Uniform}(0,1)$ and
$E\mid U=u\sim\operatorname{Bernoulli}(p(u))$.
The interval endpoints are $(0,0.20,0.35,0.60,0.75,1)$ with error
probabilities $(0.02,0.60,0.02,0.60,0.12)$ and $\alpha=0.20$.
Population retention equals the threshold. Error mass is exactly integrated over the accepted prefix. A total of 500 independent
full datasets of size 51 supply all deleted rules for unrestricted
complete-scan LEC with $c=1$. Mean reserve, signed gap, directional
bound, and absolute bound are 0.03190, 0.02537, 0.08904, and 0.12243.
The mean leave-one-out loss is $-0.00653$, compared with mean fresh-test risk $-0.00369$ from integration of the rule fitted to 50 observations. Across these same 500
full datasets, the paired difference is $-0.00284$ with Monte Carlo
standard error $0.00255$ and approximate 95\% interval
$[-0.00785,0.00217]$. Independent polynomial integration gives the
reference risk $-0.003516$, consistent with both estimates at their reported Monte Carlo precision. The individual
standard errors are $0.00206$ (LOO) and $0.00095$ (fresh-test integration).
These are computational uncertainty checks, not risk certificates.
Neither diagnostic establishes
a positive marginal risk or certifies unrestricted LEC. The directional
bound is 27.3\% below the absolute bound, and the pointwise signed
identity has maximum residual $5.6\times10^{-17}$ over these datasets.

\begin{figure*}[!tbp]
\centering\includegraphics[width=\textwidth]{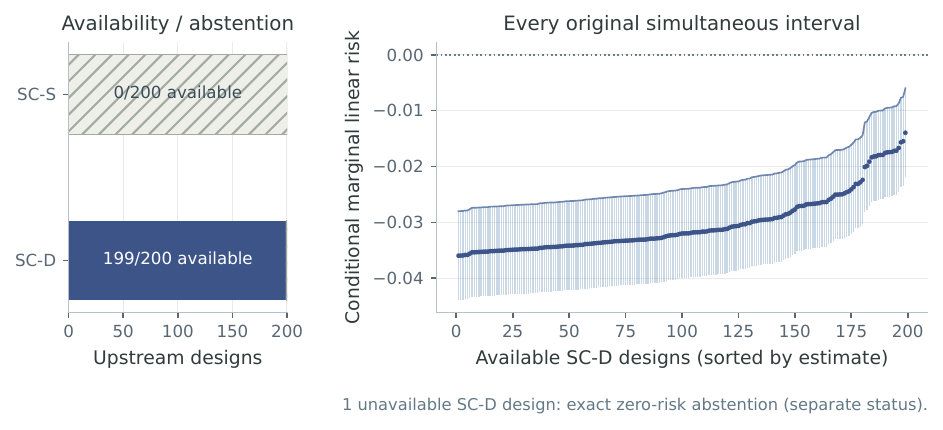}
\caption{Known-distribution evaluation with a small calibration sample. Left: available and abstaining designs. Right: conditional marginal linear risk for the 199 available SC-D designs, sorted by the estimate and shown with simultaneous Monte Carlo bounds. The remaining SC-D design and all SC-S designs abstain, giving zero linear risk and undefined SCER. Table~\ref{tab:followup-nested} reports the separate strong-core results.}
\label{fig:ttt-certified-risk}
\end{figure*}

\paragraph{Margin coverage and conditional risk.}
For this distribution, we draw 200 independent upstream datasets, each with 300 selection and 3,000 certification observations. Each design uses 20 candidates, $\delta=0.05$, and the coverage-anchor rule. Certification passes in 199/200 datasets. Simultaneous margin coverage has zero misses,
with one-sided 95\% Clopper--Pearson upper bound 1.49\%.
Each fixed design receives 500 fresh calibrations of size 300 and
exactly integrated fresh-test risk. Let $w$ be the population
linear-risk range, including abstention. With $J=200$ designs,
$B_{\rm in}=500$ inner draws, two methods, and evaluation budget
$\varepsilon=0.01$, simultaneous inner evaluation uses
\begin{equation}
\epsilon_{\rm MC}=w\sqrt{\frac{\log(4J/\varepsilon)}{2B_{\rm in}}}.
\label{eq:exp-ttt-mc-radius}
\end{equation}
An unavailable design has exactly zero risk and zero evaluation
radius. Figure~\ref{fig:ttt-certified-risk} separates these
abstentions from available designs: SC-S is unavailable throughout
this small-calibration setting, whereas all direct designs
have nonpositive upper evaluation bounds, including one abstention.

The strong-core law has endpoints $(0,0.50,0.65,0.80,0.90,1)$ and
errors $(\pi_{\rm core},0.90,0.02,0.90,0.60)$, again with $\alpha=0.20$.
At $\pi_{\rm core}=0.02$, its linear risk changes sign repeatedly and full
acceptance has error 0.298. Eleven settings each use 1,000 independent
upstream draws with 300 selection observations and margin anchors.
Coverage is checked against exact population risks for every candidate,
not just the selected anchor (Table~\ref{tab:followup-coverage}).
For $L=1$, selection-only KL bounds with $\delta=0.05$ on 20 candidates
choose the passing candidate with largest margin, or the first candidate
if none passes. Independent certification then tests that candidate.
For each core strength, a separate 200-by-500 nested evaluation
uses $n=1000$ and the same evaluation budget 0.01.
Table~\ref{tab:followup-nested} distinguishes useful retention from
zero-risk abstention: strong-core SC-S is available in 200/200 designs,
retains 51.38\%, and has nonpositive upper evaluation bounds in all
200. The nested evaluation distinguishes useful certified expansion from
zero-risk abstention; the finite-sample guarantee follows from the theorem,
rather than from the observed absence of coverage misses.

\begin{table}[!tbp]
\centering
\begingroup
\setlength{\tabcolsep}{3pt}
\renewcommand{\arraystretch}{1.08}
\footnotesize
\begin{tabular}{lrrr}
\SCtabletop
\rowcolor{SCHead}
Setting & Pass (\%) & Misses & Upper (\%) \\
\SCtablemid
Default & 100.00 & 0 & 0.30 \\
$m=100$ & 0.00 & 0 & 0.30 \\
$m=300$ & 89.80 & 1 & 0.47 \\
$m=1000$ & 100.00 & 0 & 0.30 \\
$L=1$ & 93.40 & 0 & 0.30 \\
$L=5$ & 100.00 & 2 & 0.63 \\
$L=50$ & 100.00 & 0 & 0.30 \\
$\delta=0.01$ & 100.00 & 0 & 0.30 \\
$\delta=0.10$ & 100.00 & 0 & 0.30 \\
$\pi_{\rm core}=0.10$ & 100.00 & 0 & 0.30 \\
$\pi_{\rm core}=0.19$ & 0.00 & 1 & 0.47 \\
\SCtablebottom
\end{tabular}
\endgroup

\caption{Known-law simultaneous margin coverage, 1,000 independent
draws per setting. Default: $m=3000,L=20,\delta=0.05,\pi_{\rm core}=0.02$.
Misses check every candidate, including nonpassing ones. Upper is a
pointwise one-sided binomial 95\% bound, separate from the certificate
budget and not simultaneous across settings. Rules are prespecified.}
\label{tab:followup-coverage}
\end{table}

\begin{table}[!tbp]
\centering
\begingroup
\setlength{\tabcolsep}{2.2pt}
\renewcommand{\arraystretch}{1.08}
\footnotesize
\begin{tabular}{lrrrrrr}
\SCtabletop
\rowcolor{SCHead}
$\pi_{\rm core}$ & Method & Available & \shortstack{Ret.\\(\%)} & \shortstack{$R(H)$\\$>0$} & \shortstack{$R(H)$\\$\leq0$} & Unresolved \\
\SCtablemid
\multirow{2}{*}{0.02} & SC-S & 200 & 51.38 & 0 & 200 & 0 \\
 & SC-D & 200 & 56.47 & 0 & 200 & 0 \\
\multirow{2}{*}{0.10} & SC-S & 0 & 0.00 & 0 & 200 & 0 \\
 & SC-D & 200 & 50.58 & 0 & 200 & 0 \\
\multirow{2}{*}{0.19} & SC-S & 0 & 0.00 & 0 & 200 & 0 \\
 & SC-D & 0 & 0.00 & 0 & 200 & 0 \\
\SCtablebottom
\end{tabular}
\endgroup

\caption{Known-law nested evaluation: 200 fixed designs per branch
and 500 fresh calibrations per design, with exact fresh-test integration.
Risk classifications retain simultaneous inner Monte Carlo uncertainty.
The risk columns denote confirmed positive, confirmed nonpositive,
and unresolved conditional marginal risk.
Availability is shown separately. Unavailable designs have zero risk
through abstention and undefined SCER, not useful retention.}
\label{tab:followup-nested}
\end{table}

\subsection{Computational Cost}
\label{app:followup-cost}
We report median CPU times over seven repetitions after warmup for sorting, prefix scanning, the complete calibration scan, and 20-candidate KL certification in a single CPU process. Table~\ref{tab:followup-cpu} gives the measurements.
At $n=10^5$, total scanning takes 18.40 ms and certification 21.24 ms.
Naive leave-one-out diagnostics are excluded from deployment cost.
These CPU measurements exclude model inference and were taken before
the conservative numerical checks described in
Appendix~\ref{app:implementation}; they therefore do not measure the runtime
of the final implementation.
Because prefix loss is non-monotonic, binary search can return a
different threshold. Every reported experiment uses complete scanning,
which is required to recover the largest feasible prefix without a
monotonicity assumption.

\begin{table}[!tbp]
\centering
\begingroup
\setlength{\tabcolsep}{3pt}
\renewcommand{\arraystretch}{1.08}
\footnotesize
\begin{tabular}{lrrrr}
\SCtabletop
\rowcolor{SCHead}
$n$ & Sort & Scan & Total & Cert. \\
\SCtablemid
100 & 0.005 & 0.022 & 0.024 & 2.979 \\
1000 & 0.078 & 0.030 & 0.103 & 3.142 \\
10000 & 1.230 & 0.097 & 1.515 & 4.424 \\
100000 & 17.254 & 0.803 & 18.399 & 21.241 \\
\SCtablebottom
\end{tabular}
\endgroup

\caption{Single-process CPU medians (ms). Total time includes sorting and scanning. Separately measured component medians need not sum to the total median. Model inference and leave-one-out diagnostics are excluded. These timings exclude the final conservative numerical checks.}
\label{tab:followup-cpu}
\end{table}

\subsection{Routing Application}
\label{app:followup-routing}
\begin{figure*}[!tbp]
\centering\includegraphics[width=\textwidth]{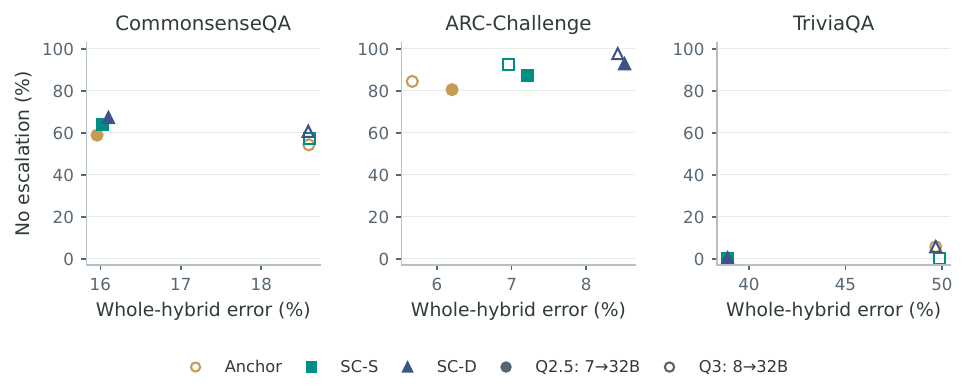}
\caption{Prespecified routing pairs, $\alpha=0.20$, margin/KL,
over all 100 repetitions including unavailable designs. Colors encode
selectors and shapes also encode methods. Filled/open markers encode
the Qwen2.5/Qwen3 pairs. No-escalation rate (cheap-model answers)
and overall
hybrid error are empirical offline-replay quantities. The cheap-branch
SCER certificate does not transfer to the plotted whole-system error.}
\label{fig:routing}
\end{figure*}

Six prespecified pairs route Qwen2.5-7B to Qwen2.5-32B and Qwen3-8B
to Qwen3-32B on each dataset. We evaluate these pairs offline using the fitted margin/KL policies over the 100 paired repetitions. Each query is first answered by the cheaper model. Accepted answers are retained, and rejected queries are sent to the stronger model. Figure~\ref{fig:routing} reports results at $\alpha=0.20$. On CommonsenseQA, Qwen2.5 SC-D calls the strong
model on 32.77\% of questions, with cheap-branch SCER 8.03\% and
overall hybrid error 16.10\%.
The certificate applies only to answers retained from the cheaper model, with the upstream confidence qualification. Whole-system error is an empirical routing outcome without a corresponding system-level risk guarantee. The offline analysis does not measure online latency or savings.

\subsection{A Non-Monotonic Calibration Counterexample}
\label{app:lec-counterexample}
We consider an exactly specified four-band i.i.d. population law,
distinct from the mechanism diagnostic in
Appendix~\ref{app:followup-certification}.
Draw i.i.d. $U\sim\operatorname{Uniform}(0,1)$ and set
\[
E=\mathbf1\{U\in[0.10,0.20)\cup[0.60,1]\}.
\]
Write $p(u)=\Pr(E=1\mid U=u)$ for this error law.
A fixed classifier predicting zero with feature $X=U$, label $Y=E$,
and score $U$ realizes the law without using test labels.
The ranking contains a perfect but limited safe core and a second
correct-answer interval. The population linear-risk slopes are
$(-0.20,0.80,-0.20,0.80)$ on intervals with endpoints
$(0,0.10,0.20,0.60,1)$. Its nonzero roots are $0.125$, $0.50$, and
$0.625$, so risk changes sign repeatedly.

Use $n=25$, $\alpha=0.20$, and the original LEC Algorithm~2
of \citet{wang2026lec}: inspect every observed score and return the
largest whose cumulative loss is at most $-1$, or abstain if none
exists. This is $A_1$ in Eq.~\eqref{eq:empirical-linear-sum}, with the published
buffer and fallback. The law is i.i.d., scores are almost surely distinct,
and the predictor and score map are fixed. It therefore satisfies the
published theorem's exchangeability assumption and fixed-model setup.

\paragraph{Exact population calculation.}
Let $\boldsymbol V=(V_1,V_2,V_3,V_4)$ count calibration points in
the four intervals, with realizations in
\[
\mathcal V_{25}=\left\{\boldsymbol v\in\mathbb N_0^4:
\sum_{b=1}^4v_b=25\right\}.
\]
Their widths are
$\boldsymbol w=(1/10,1/10,2/5,2/5)$ and errors
$\boldsymbol e=(0,1,0,1)$. Each count vector has probability
\[
\pi(\boldsymbol v)=\frac{25!}{\prod_bv_b!}\prod_bw_b^{v_b}.
\]
Conditional on these counts, the ordered errors are
$0^{v_1}1^{v_2}0^{v_3}1^{v_4}$. Let $K(\boldsymbol v)$ be their
largest feasible prefix length, or zero for abstention. When the
selected observation is the $j$th point in interval $b$, its conditional
mean threshold is
\[
\ell_b+w_b\frac{j}{v_b+1},
\]
where $\ell_b$ is that interval's left endpoint. Population risk is
linear within each interval. Therefore its conditional expectation is
\[
R_P(\ell_b)+(e_b-\alpha)w_b\frac{j}{v_b+1},
\]
with $R_P(t)=\int_0^t(p(u)-\alpha)\,du$, and zero under abstention.
Write $\rho(\boldsymbol v)=\mathbb E[R_P(\widehat\lambda)\mid
\boldsymbol V=\boldsymbol v]$ for this conditional risk.
Summing its rational values over all $\binom{28}{3}=3276$ count vectors gives
\begin{equation}
\begin{aligned}
\mathcal R_{25}(A_1)
&=\sum_{\boldsymbol v\in\mathcal V_{25}}
\pi(\boldsymbol v)\rho(\boldsymbol v)\\
&\approx2.078\times10^{-3}>0.
\end{aligned}
\label{eq:lec-exact-counterexample}
\end{equation}
We evaluate the finite sum in Eq.~\eqref{eq:lec-exact-counterexample}
using exact rational arithmetic; the displayed decimal is an approximation.
The exact marginal retention is approximately $0.259713393$, and
SCER is approximately $0.208001258>0.20$. Non-abstention has probability
approximately $0.459875794$. All abstentions enter the risk and retention averages. This is a population counterexample, not an
empirical test-set exceedance.

An independent enumeration of the $N=26$ full-data count vectors
($\binom{29}{3}=3654$ cases) gives
$\mathbb E[H_1]\approx0.022722073$ and
$\Delta_1\approx0.024800107$. Their exact rational difference equals
Eq.~\eqref{eq:lec-exact-counterexample}. Thus the signed gap exceeds the reserve in Figure~\ref{fig:lec-exact-counterexample}. The violation follows from the exact population calculation, rather than from an upper bound that exceeds the reserve.

\paragraph{Scope of the original guarantee.}
LEC's Theorem~3.1 claims marginal control under exchangeability.
Its Appendix~A.1, Eq.~(17), transfers the loss at a calibration-fitted
threshold to an average including calibration observations. Exchangeability
alone does not justify that equality: fitting privileges calibration
points relative to the test point. The pointwise loss is non-monotonic,
so the monotone CRC argument invoked there does not supply the missing
transfer. The empty-feasible-set branch also has zero calibration loss,
rather than the $-1$ reserve used on feasible samples. Our calculation
explicitly includes that branch.

The published Algorithm~2 specifies observed-score candidates, which
we follow above. Reading its Eq.~(7) literally as a continuum supremum
introduces a nonattainment issue: if the last feasible prefix has size
$k<n$, the supremum is $U_{(k+1)}$, which need not itself be feasible
under the $\le$ convention. A feasible full prefix gives acceptance of
the entire support. Exact integration with this literal supremum and
the same failure abstention also gives positive risk (approximately $0.014350$)
and SCER approximately $0.250324$. The counterexample therefore does not
depend on replacing that supremum by a more adverse search rule.
This counterexample concerns the published threshold-selection rule. It does not invalidate the ratio-to-linear identity or imply that the rule fails on every distribution.

\paragraph{Discovery and independent confirmation.}
We first examined 33 distribution and sample-size settings using multinomial integration, without QA test outcomes. After fixing the distribution, $n$, and selection rule, we verified the counterexample by exact rational enumeration with an independent prefix-scan implementation. Separately, 500,000 independent
calibration draws and exact fresh-test integration give mean risk
$0.002092$. Simultaneous 99\% Hoeffding bounds for risk and retention
give risk interval $[0.001308,0.002876]$ (rounded outward) and SCER interval
$[0.204995,0.211183]$. The bounds use ranges $[-0.02,0.30]$ for risk
and $[0,1]$ for retention. This separate Monte Carlo calculation checks the numerical result. Exact enumeration establishes the counterexample.

\paragraph{SC-LEC on the same law.}
The externally fixed anchor $\lambda_0=0.10$ has true margin
$\eta=0.02$ and retention $0.10$. With this known valid margin and
the same $n=25$, the parameter rules in Appendix~\ref{app:automatic-parameters} give SC-D $c=h=0$. SC-S is unavailable and abstains.
Table~\ref{tab:lec-counterexample-methods} shows the complete comparison.
Neither certificate permits expansion at this small-$n$ setting. SC-D retains the anchor and SC-S has zero retention.

\begin{table}[t]
\centering\footnotesize
\begin{tabular}{lrrr}
\SCtabletop
\rowcolor{SCHead}
Method & Ret. (\%) & SCER (\%) & Linear risk\\
\SCtablemid
LEC & 25.9713 & 20.8001 & $0.002078$\\
Anchor & 10 & 0 & $-0.02$\\
SC-S & 0 & -- & $0$\\
SC-D & 10 & 0 & $-0.02$\\
\SCtablebottom
\end{tabular}
\caption{Exact-law comparison with a known valid margin and the
same final-calibration size. SC-S's zero risk comes from fallback
abstention, with undefined SCER. The anchor and SC-D use population
margin information. The comparison therefore does not match the information available to each method.}
\label{tab:lec-counterexample-methods}
\end{table}

To evaluate independent certification, we use this single external anchor ($L=1$), $m=3000$, $\delta=0.05$,
and 5000 independent upstream draws. Certification passes in 4918/5000
draws, with no observed margin noncoverage. Every observed available
SC-D design has $h=0$ and SC-S remains unavailable. Including failures,
Anchor and SC-D have mean retention $9.836\%$ and mean conditional
linear risk $-0.019672$. With estimated margins, the guarantee retains its upstream confidence qualification. The observed certification frequency does not convert it into unconditional exact SCER control.
LEC uses 25 labels, Anchor 3000, and SC-S/SC-D 3025. No extra selection
labels are used. This label imbalance is separate from the risk-transfer
violation established by exact enumeration.

\subsection{LEC Target Attainment on QA Benchmarks}
\label{app:lec-real-qa-diagnostics}
We analyze fixed prediction records, fitted policies, and question-role
assignments without refitting them. The analysis covers all 60 primary
dataset/model/target configurations (three datasets, four Qwen models, five targets, 100 repetitions each). Aggregate
SCER is always total accepted errors divided by total accepted predictions,
with abstentions included. It is not a mean of per-run ratios.

Table~\ref{tab:lec-diagnostic-summary} summarizes the complete primary
grid. Table~\ref{tab:cross-dataset} reports retention, empirical SCER,
and unavailable SC-S/SC-D outcomes at $\alpha=0.20$;
Figure~\ref{fig:full-kl-sweep} gives retention across all targets.
The intervals jointly resample the 100 repetition assignments for paired
methods, using 10,000 bootstrap samples. These pointwise intervals measure split-to-split variability conditional on the fixed benchmark pool. They do not quantify uncertainty for a new question population or the probability of certificate failure. Repeated appearances of the same question are not
treated as independent binomial trials. Undefined zero-acceptance SCER
and certificate unavailability remain explicit.

LEC's aggregate empirical SCER exceeds its target in 21/60 primary
configurations. Four pointwise lower interval endpoints also exceed
their targets. These are descriptive counts over a reported grid,
without a simultaneous population claim. They are distinct from the
exact known-law violation above. All repetitions, including those with no
accepted answers, enter these aggregate results; no minimum-acceptance
filter is applied.

\begin{table}[t]
\centering\footnotesize
\begin{tabular}{lrrr}
\SCtabletop
\rowcolor{SCHead}
Task & Cells & \shortstack{Empirical\\exceedances} & \shortstack{Lower CI\\above $\alpha$}\\
\SCtablemid
CommonsenseQA & 20 & 6 & 0 \\
ARC-Challenge & 20 & 2 & 0 \\
TriviaQA & 20 & 13 & 4 \\
\SCtablebottom
\end{tabular}
\caption{Descriptive LEC target attainment over the complete primary
grid. Each task has four models and five targets. The last column
counts pointwise split-bootstrap intervals whose lower endpoint exceeds
the target. All primary configurations enter these counts; this is not a
simultaneous population-validity test.}
\label{tab:lec-diagnostic-summary}
\end{table}

\section{Implementation Details}
\label{app:implementation}
This appendix specifies the calibration algorithm, parameter calculations, and inference protocol. The anchor, margin, buffer, and radius are fixed before final calibration; the final scan selects a threshold within that certified window.

\subsection{End-to-End SC-LEC Algorithm}
\label{app:algorithm}
Algorithm~\ref{alg:sclec} takes independent data roles and a prespecified
design policy. The main protocol maximizes certified margin. The coverage policy is a separate prespecified variant. Candidate levels,
buffer fractions, tie order, and the full-abstention fallback are fixed
before calibration.

\begin{algorithm}[!b]
\caption{Independently Certified SC-LEC}
\label{alg:sclec}
\small
\begin{algorithmic}[1]
\Require Independent $D_{\rm sel},D_{\rm cert},D_{\rm cal}$; $\alpha,\delta$;
\Statex \hspace{\algorithmicindent} fixed score/tie order, anchor policy, quantile levels,
\Statex \hspace{\algorithmicindent} buffer-fraction grid, and branch $\tau\in\{\text{SC-S},\text{SC-D}\}$
\Ensure Certified design and $\widehat\lambda$, or full abstention $\bot$
\State $m\gets|D_{\rm cert}|$; $n\gets|D_{\rm cal}|$; $N\gets n+1$
\State $\Lambda_0\gets$ distinct selection-score quantiles; $L\gets|\Lambda_0|$
\For{$j=1,\ldots,L$}
  \State $\overline Y_j\gets m^{-1}\sum_{W\in D_{\rm cert}}[g_\alpha(W;\lambda_j)+\alpha]$
  \State $u_j\gets$ KL upper bound in Eq.~\eqref{eq:certification-average}
\EndFor
\State $\mathcal P\gets\{j:u_j<\alpha\}$
\If{$\mathcal P=\varnothing$}
  \State \Return $\bot$ \Comment{No certified anchor}
\EndIf
\State $j^\star\gets$ prespecified anchor policy applied to $\mathcal P$
\State $\lambda_0\gets\lambda_{j^\star}$; $\widehat\eta\gets\alpha-u_{j^\star}$
\If{$\tau=\text{SC-D}$}
  \State $c\gets0$; $h\gets h_{\rm D}(0)$ from Appendix~\ref{app:automatic-parameters}
\Else
  \State Evaluate valid $(c,h_{\rm S}(c))$ on the fixed buffer grid
  \If{no valid pair exists}
    \State \Return $\bot$ \Comment{Certificate unavailable}
  \EndIf
  \State Choose largest $h$, breaking ties by smaller $c$
\EndIf
\State Freeze $(\lambda_0,\widehat\eta,c,h)$ before inspecting calibration values
\State Sort $D_{\rm cal}$ by the fixed order; $C_0\gets0$
\For{$j=1,\ldots,n$}
  \State $C_j\gets C_{j-1}+E_{(j)}-\alpha$
\EndFor
\State $r_0\gets\#\{i:U_i\le\lambda_0\}$; $h'\gets\min\{h,n-r_0\}$
\State $\widehat\lambda\gets\bot$
\For{$j=r_0,\ldots,r_0+h'$} \Comment{Inspect every candidate}
  \If{$C_j\le-c$}
    \If{$j=r_0$}
      \State $\widehat\lambda\gets\lambda_0$
    \Else
      \State $\widehat\lambda\gets U_{(j)}$
    \EndIf
  \EndIf
\EndFor
\State \Return $(\lambda_0,\widehat\eta,c,h,\widehat\lambda)$
\end{algorithmic}
\end{algorithm}

The final scan uses calibration scores and error labels only. It examines every permitted candidate, including candidates beyond an empirically infeasible anchor.

\subsection{Automatic Certificate Parameter Selection}
\label{app:automatic-parameters}
The following calculations implement Theorems~\ref{thm:explicit-scer}
and~\ref{thm:direct-anchor} without estimating a stability expectation.

\paragraph{Data protocol.}
We fix $m$, $n$, the candidate-grid construction, and $\delta$ before observing certification outcomes. A concrete grid consists of
distinct empirical quantiles of $U$ on $D_{\rm sel}$ at prespecified
levels. We remove duplicates and denote the resulting grid size by $L$.
The confidence calculation conditions on the entire grid,
so its random values and size incur no additional penalty.
If a single anchor is selected using only $D_{\rm sel}$ and then
certified on independent data, $L=1$ suffices.
If multiple anchors are inspected on $D_{\rm cert}$, the simultaneous
correction is required. Candidates generated from certification
scores are not covered by this finite-grid argument.
The model and uncertainty estimator must remain unchanged after
their predictions are used for certification.

In an i.i.d. experiment, assign observations to disjoint roles using
a random split independent of their values. There is no universally
valid optimal split proportion: certification needs enough samples
to resolve the margin, while $n$ determines the expansion bound.
We choose these sizes using prior knowledge or selection data, then fix them before certification. The evaluation set may contain
many i.i.d. test points, but their empirical SCER is an estimate of
the population target, not an additional exact finite-test guarantee.
The protocol keeps certification and final-calibration observations disjoint. Final calibration and test outcomes are not used to tune $\lambda_0,c,h$. Repeated certification attempts or adaptive stopping
require a separately valid error-spending or sequential procedure.

\paragraph{Scalar confidence-bound computation.}
For $y=\overline Y_j\in[0,1)$, solve
\[
\operatorname{kl}(y\Vert u)=\log(L/\delta)/m,
\qquad u\in(y,1).
\]
The root is unique, because KL is increasing in $u$ above $y$,
equals zero at $u=y$, and diverges as $u\uparrow1$.
For $y=0$, the closed form is
\[
u=1-(\delta/L)^{1/m}.
\]
For $y=1$, use $u=1$ and reject the candidate.
Floating-point bisection on $[y,1]$ proposes candidate endpoints but does not certify which side of the exact root they lie on. We verify each bracket update using outward-rounded residual intervals for the represented inputs:
a lower endpoint requires a strictly negative residual upper bound,
and an interior upper endpoint requires a nonnegative residual lower
bound. The analytic endpoint one remains a valid upper bound.
The returned confidence endpoint is therefore verified, including
when the zero-mean closed form supplies the proposal. Certified margins are rounded downward. Endpoint-safe evaluation uses
$0\log0=0$ and the appropriate infinite limits.
The lowest possible upper bound occurs at $y=0$. Consequently,
even the most favorable certification sample cannot pass unless
\[
m>\frac{\log(L/\delta)}{-\log(1-\alpha)}.
\]
This is a necessary feasibility check, not a sufficient sample-size
guarantee for an arbitrary population anchor.

\paragraph{Direct parameter choice.}
Write $\eta=\widehat\eta$ in this paragraph. For any
$0\le c<n\eta$, the largest allowed radius under the direct
certificate is
\[
h_{\rm D}(c)=\min\left\{n,
\left\lfloor\frac{N\eta(1-q_{n,c})}{1-\alpha}\right\rfloor
\right\}.
\]
To verify that $c=0$ maximizes this radius, set $p=\alpha-\eta$
and $a(c)=\alpha-c/n>p$. Then
\[
\frac{\partial}{\partial a}\operatorname{kl}(a\Vert p)
=\log\frac{a(1-p)}{p(1-a)}>0.
\]
Increasing $c$ decreases $a(c)$, decreases the KL exponent, and
increases $q_{n,c}$. Hence $h_{\rm D}(c)$ is nonincreasing.
For a fixed candidate window, increasing $c$ also shrinks the
empirically feasible set. We use $c=0$ because it gives the largest certified radius and the least restrictive empirical feasibility condition for a fixed window. A prespecified positive buffer is also supported through $h_{\rm D}(c)$.

\paragraph{Stability parameter choice.}
An executable grid is $\rho\in\{0,0.05,\ldots,0.95\}$, with
$c=\rho n\eta$. Discard any $c$ with $c+\alpha\ge N\eta$.
For a remaining buffer, let
\[
a_c=\frac{c(1-q_{N,c})}{M_\alpha}
-N\widetilde q_{N,c}.
\]
If $a_c<0$, no radius, including zero, is certified by
Theorem~\ref{thm:explicit-scer}. Otherwise, $\widetilde q_{N,c}<1$
and the largest certified radius is
\[
h_{\rm S}(c)=\min\left\{n,
\left\lfloor\frac{a_c}{1-\widetilde q_{N,c}}\right\rfloor\right\}.
\]
Indeed, Eq.~\eqref{eq:explicit-condition-tight} is equivalent to
\[
h(1-\widetilde q_{N,c})\le a_c.
\]
We maximize $h_{\rm S}(c)$ over the retained buffers and break ties by smaller $c$. This choice maximizes the certified radius over the specified grid rather than optimizing population retention over all designs.
Use outward upper tail endpoints to form conservative radius bounds,
then floor and recheck the original certificate with exact rational
arithmetic. Ordinary floating-point tail values are diagnostic only and cannot certify a radius at a rounding boundary. We fix the chosen certificate, $\lambda_0,u_{j^\star},\widehat\eta,c,h$, split
sizes, and $\delta$ before processing final calibration data.

\paragraph{Acceptance ordering under a shared design.}
\label{app:default-design-ordering}
Consider the KL parameter rules with the same fixed anchor, margin
$\eta$, sample size $n$, and score order. If $\alpha\le1/3$, the default
SC-D design ($c_{\rm D}=0$, largest certified radius) accepts at least
as much as any SC-S-certified design, pointwise in the calibration
and test observations. This orders acceptance, not population risk.

To see this, write $M=1-\alpha$, so $2\eta<M$. For any valid SC-S
pair $(c,h)$, its assumptions imply $c<n\eta$.
Moreover $q_{n,c}\le\widetilde q_{N,c}$: the two Bernoulli-Chernoff
problems have the same total threshold $n\alpha-c$, while adding
one nonnegative Bernoulli summand increases the Chernoff bound.
Formally, with $p=\alpha-\eta$ and $a(t)=1-p+pe^t\ge1$ for $t>0$,
compare the infima of
$e^{-t(n\alpha-c)}a(t)^n$ and
$e^{-t(n\alpha-c)}a(t)^N$.
SC-S certification gives $Mh\le cp_{N,c}<N\eta$, hence
$M(N-h)>N(M-\eta)>N\eta$. Therefore
\[
\begin{aligned}
Mh&\le cp_{N,c}-M(N-h)\widetilde q_{N,c}\\
&\le c-N\eta\widetilde q_{N,c}\\
&\le N\eta(1-q_{n,c}).
\end{aligned}
\]
Thus SC-D also certifies radius $h$ at buffer $c$. Reducing the direct
buffer to zero only decreases $q_{n,c}$, so its largest certified
radius is at least $h$. Its candidate set contains SC-S's and its
empirical constraint $T_n\le0$ is weaker than $T_n\le-c$.
Consequently its largest feasible threshold is no smaller, with
abstention ordered below every threshold. An unavailable SC-S design
also satisfies this ordering through its abstention fallback.
This inclusion applies to our primary targets $0.05$--$0.30$ under the shared-anchor
protocol. It does not order arbitrary buffer/radius policies, different anchors, or realized selective risks.

\paragraph{Exact prefix indexing.}
With sorted calibration scores $U_{(1)}<\cdots<U_{(n)}$, define
\begin{equation}
C_0=0,\qquad C_j=\sum_{r=1}^j(E_{(r)}-\alpha).
\label{eq:prefix-risk}
\end{equation}
The anchor's empirical loss is $C_{r_0}$ even if $\lambda_0$ is not an
observed calibration score. Writing $h'=\min\{h,n-r_0\}$,
\begin{equation}
\begin{aligned}
\mathcal J_h&=\{r_0,r_0+1,\ldots,r_0+h'\},\\
\widehat j_{c,h}&=\max\{j\in\mathcal J_h:C_j\le-c\},
\end{aligned}
\label{eq:local-index-set}
\end{equation}
with an empty feasible set giving abstention. Otherwise
\begin{equation}
\widehat\lambda=
\begin{cases}
\lambda_0,&\widehat j_{c,h}=r_0,\\
U_{(\widehat j_{c,h})},&\widehat j_{c,h}>r_0.
\end{cases}
\label{eq:local-threshold}
\end{equation}
The unrestricted rule instead scans $j\in\{1,\ldots,n\}$ and outputs
the largest feasible observed score. Sorting costs $O(n\log n)$,
prefix computation $O(n)$, and the local scan at most $h+1$ checks.

\paragraph{Nested windows.}
\label{app:nested-windows}
Fix the anchor, buffer, final sample, and strict score/tie order.
For integers $h_1\le h_2$, clipping at $n-r_0$ preserves
$\mathcal J_{h_1}\subseteq\mathcal J_{h_2}$ in
Eq.~\eqref{eq:local-index-set}. Feasibility $C_j\le-c$ does not depend
on the radius, so the feasible sets are nested too. If the smaller
set is empty, its output is $\bot$, which is ordered below every threshold. Otherwise, the larger set is nonempty and its maximum is no smaller.
This includes $h=0$, an infeasible anchor followed by a feasible
extension, and $r_0=n$, when all windows coincide. The output map in
Eq.~\eqref{eq:local-threshold} preserves this order: rank $r_0$ maps
to $\lambda_0$ and higher ranks to strictly larger score/tie pairs.
Consequently, for every evaluation observation $w$,
\[
S(w;A_{c,h_1}(D_n))\le S(w;A_{c,h_2}(D_n)).
\]
Calibration accepted counts and, integrating over a fresh observation,
population retention for these fixed fitted policies are nondecreasing.
This population comparison integrates the pointwise acceptance inclusion over the same deployment score and tie-key distribution, rather than inferring it from calibration counts alone. The argument
also holds conditional on upstream certification, and averaging over
the common final sample preserves the inequality. Raw score ties are
resolved by independent continuous keys, giving a strict lexicographic
order almost surely. Neither selective error nor linear risk need be
monotone. Changing the anchor or buffer invalidates this particular
nested-set comparison.

\paragraph{Boundary cases and ties.}
If $r_0=n$, no above-anchor candidate exists, irrespective of $h$.
If $r_0=0$, the anchor still represents a population threshold: $C_0=0$
does not mean that it has zero population retention. At $c=0$ this
candidate is feasible even when no calibration point is accepted.
At $h=0$ the algorithm returns the anchor or abstention, as analyzed
in Appendix~\ref{app:proof-retention}. A missing certified anchor and
an unavailable SC-S design are distinct outcomes, although both use
full abstention under the main protocol. Genuine score ties use the
independent continuous key described in Appendix~\ref{app:server-scores}. The same lexicographic comparison defines quantiles, ranks, and deployment.
No numerical jitter or outcome-dependent ordering is introduced.

\paragraph{Numerical conventions.}
Prefix feasibility uses integer error counts and an exact scaled comparison
with the represented buffer. KL inversion returns an upper endpoint verified
by outward-rounded residual intervals. Certified margins are rounded downward.
Tail probabilities are enclosed outward, and radius selection rechecks the
original certificate using exact rational arithmetic at the upper tail
endpoints. We round reported risk bounds upward and retention lower bounds downward.
These conventions conservatively implement the stated sufficient conditions.

\subsection{Exact Prompt Templates}
\label{app:prompts}
The following templates specify the user messages used for inference. Bracketed fields denote
dataset text, not additional prompt instructions.

\paragraph{CommonsenseQA and ARC-Challenge.}
For each option, the implementation appends its label, a colon, a space,
and its original text, preserving option order. The message is
\par\noindent\begin{minipage}{\linewidth}
\begin{quote}
\raggedright
\ttfamily
[question]\\
A: [first option]\\
B: [second option]\\
\normalfont\emph{(one line per remaining option)}\\
\ttfamily
Answer with the option label only.
\end{quote}
\end{minipage}\par
The parenthetical line above denotes repeated template structure and is
not sent to the model. CommonsenseQA has five options. ARC uses the
question's available options after mapping labels to A/B/\ldots.
Prediction scores the fixed label completions instead of sampling a
free-form option answer. Thus the instruction and the label-likelihood
policy are both part of reproducing the predictor.

\paragraph{Closed-book TriviaQA.}
The user message is exactly
\begin{quote}
\raggedright
\ttfamily
[question]\\
Answer briefly with the answer only.
\end{quote}
No retrieved context, aliases, answer key, or demonstration is included.
The answer is generated greedily with at most 36 new tokens.

\paragraph{Model-specific serialization.}
Both model families receive a single user message through their official
tokenizer chat template with a generation prompt. No explicit system
message is supplied by the inference code. Template-inserted defaults are those of the pinned tokenizer. Qwen3 thinking is disabled with
\texttt{enable\_thinking=False}. Both families receive the same user-facing text, serialized through their respective official chat templates with model-specific delimiters.

\subsection{Prediction and Uncertainty Calculations}
\label{app:prediction-example}
This section defines the uncertainty scores and error indicators used in the experiments.

\paragraph{Multiple choice.}
For a dataset question $q$ with ordered choices $(a_1,\ldots,a_K)$,
let $\ell_k$ be the summed log-likelihood of the fixed completion
tokens for label $k$. Normalize only across the available labels:
\[
p_k=\frac{\exp(\ell_k)}{\sum_{s=1}^K\exp(\ell_s)},\qquad
\widehat k=\arg\max_k p_k.
\]
The predictor answers $a_{\widehat k}$. Its primary uncertainty and
error indicator are
\[
U=-\frac{\sum_{k=1}^K p_k\log p_k}{\log K},\qquad
E=\mathbf1\{\widehat k\ne k_{\rm gold}\}.
\]
These probabilities normalize label likelihoods, rather than the
full vocabulary, and need not themselves be calibrated.

\paragraph{TriviaQA.}
Let $t_1,\ldots,t_T$ be the generated answer tokens, excluding terminal
EOS, and let $a$ be the decoded answer. The model is teacher-forced on
that same generated sequence to obtain
\[
U=-\frac1T\sum_{r=1}^T\log p(t_r\mid q,t_{<r}).
\]
For the benchmark alias set $\mathcal A(q)$, normalized exact-match error is
\[
E=\mathbf1\{\operatorname{norm}(a)\notin
\{\operatorname{norm}(b):b\in\mathcal A(q)\}\}.
\]
Normalization follows the specified article/punctuation and whitespace
rules. Aliases determine correctness only. They do not enter generation
or the NLL. Empty output remains an error with fixed high uncertainty,
as specified in Appendix~\ref{app:server-scores}.

\subsection{Worked Design and Calibration Scan}
\label{app:calibration-example}
We illustrate the parameter choices using the first repetition of CommonsenseQA with Qwen2.5-32B-Instruct, primary option entropy, and $\alpha=0.20$. Table~\ref{tab:worked-design} reports the margin/KL design parameters, showing how the certificates determine different windows from the same anchor.
The threshold is a lexicographic pair $(U,V)$ because the fixed rule
attaches an independent tie key.

\begin{table}[tp]
\centering
\setlength{\tabcolsep}{4pt}
\begin{tabular}{lll}
\SCtabletop
\rowcolor{SCHead}
Quantity & SC-S & SC-D\\
\SCtablemid
$\lambda_0$: score & \multicolumn{2}{c}{$\approx6.25\times10^{-5}$}\\
$\widehat\eta$ & \multicolumn{2}{c}{$\approx0.0806$}\\
Certification $m$ & \multicolumn{2}{c}{3834}\\
Calibration $n$ & \multicolumn{2}{c}{2738}\\
$L$, $\delta$ & \multicolumn{2}{c}{20, 0.05}\\
$c$ & $\approx154.39$ & $0$\\
$h$ & $190$ & $275$\\
\SCtablebottom
\end{tabular}
\caption{Example calibrated designs from the first repetition on CommonsenseQA Qwen2.5-32B,
primary entropy, $\alpha=0.20$, margin/KL. The branches share the
certified anchor. Parameters are rounded for display. Ties use the prespecified key.}
\label{tab:worked-design}
\end{table}

The subsequent scan is a \emph{symbolic illustration} using the numerical design parameters above. Let
$B_0=\sum_{r\le r_0}E_{(r)}$ be the number of anchor errors and
$e_s=E_{(r_0+s)}$ the next ordered errors. Every candidate has
\[
C_{r_0+k}=B_0-\alpha r_0+\sum_{s=1}^k e_s-\alpha k,
\qquad 0\le k\le h'.
\]
Table~\ref{tab:symbolic-prefix} gives the corresponding symbolic checks.

\begin{table}[tp]
\centering
\begin{tabular}{ll}
\SCtabletop
\rowcolor{SCHead}
Candidate rank & Prefix linear loss\\
\SCtablemid
$r_0$ & $B_0-\alpha r_0$\\
$r_0+1$ & $C_{r_0}+e_1-\alpha$\\
$r_0+k$ & $C_{r_0}+\sum_{s=1}^k e_s-\alpha k$\\
$r_0+h'$ & $C_{r_0}+\sum_{s=1}^{h'}e_s-\alpha h'$\\
\SCtablebottom
\end{tabular}
\caption{Symbolic prefix scan, with $h'=\min\{h,n-r_0\}$.
Each row tests $C_j\le-c$. The corresponding threshold is $\lambda_0$ for $j=r_0$
and $U_{(j)}$ otherwise. Positive-expansion rows require $h'>0$.
}
\label{tab:symbolic-prefix}
\end{table}

SC-D uses $C_j\le0$, whereas SC-S uses $C_j\le-c$ with its chosen buffer. Let $k^\star$ be the largest feasible $k$. Each rule returns the corresponding threshold:
the anchor for $k^\star=0$,
$U_{(r_0+k^\star)}$ for $k^\star>0$, or abstention if none passes.
Correct answers lower prefix loss and errors raise it, so infeasibility
cannot terminate the scan.

\section{Additional Related Work}
\label{app:additional-related-work}

\paragraph{Selective QA and learned abstention.}
Selective prediction trades coverage for accuracy by allowing a model to
abstain~\citep{el2010foundations}. In QA, the usefulness of abstention
depends on whether confidence identifies questions the model is likely to
answer correctly. \citet{kamath2020selective} study selective QA under
domain shift and train an error-predicting calibrator using in-domain and
out-of-domain examples. Their setting motivates studying abstention beyond
raw model probabilities, although our guarantees assume the stated
exchangeability or i.i.d. conditions. Training can also change a model's
refusal behavior: R-Tuning constructs refusal-aware instruction data so
that LLMs learn when to answer and when to decline~\citep{zhang2024rtuning}.
Similarly, SelectiveNet jointly learns prediction and
rejection~\citep{geifman2019selectivenet}, whereas SGR calibrates a selection
rule with high probability~\citep{geifman2017selective}. SC-LEC calibrates
the threshold of a fixed predictor and uncertainty score, keeping this
statistical task separate from learning the answer or refusal policy.

\paragraph{Confidence estimation and calibration.}
Probability calibration asks whether confidence values reflect observed
correctness frequencies. Temperature scaling is a standard post-hoc
approach to this problem~\citep{guo2017calibration}. For language models,
\citet{kadavath2022language} study self-evaluation of proposed answers and
prediction of whether a question can be answered. \citet{tian2023just}
compare methods for eliciting confidence from models fine-tuned with human
feedback, while \citet{xiong2024uncertainty} evaluate prompting, sampling,
and aggregation for confidence calibration and failure prediction.
These studies concern the information in confidence scores and their
empirical calibration. The finite-sample risk of a threshold selected
using calibration data is a separate statistical question.
Other signals compare generated responses: SelfCheckGPT uses consistency
across samples to detect potentially nonfactual content without an external
database~\citep{manakul2023selfcheckgpt}, and semantic uncertainty groups
equivalent answers~\citep{kuhn2023semantic,farquhar2024semantic}.
Such signals can inform the score supplied to a calibration procedure.
Their ability to rank errors does not by itself certify the selected
answers' error rate.

\paragraph{Conformal guarantees for generation and selection.}
Conformal prediction controls prediction-set
coverage~\citep{angelopoulos2023conformal}, but the set or error event being
controlled matters in language generation. Conformal Language Modeling
calibrates sampling and rejection to produce a set containing at least one
acceptable response, with high probability over calibration data of meeting
the specified test-coverage target~\citep{quach2024conformal}. This controls
an event about the output set, rather than the error rate among individually
accepted answers. \citet{mohri2024factuality} instead calibrate a back-off
rule that reduces output specificity to obtain entailment-based factuality
guarantees. Their marginal guarantee is defined relative to a reference
and an entailment oracle, and changes the content of the output rather than
only selecting whether to return a fixed answer.
ConU and SConU also study conformal uncertainty and selection for language
models~\citep{wang2024conu,wang2025sconu}. For batch selection,
\citet{jin2023selection} combine conformal p-values with multiple testing
to control false discovery rate under their stated assumptions.
This is an expectation of a random error fraction, as is the batch target
in Conformal Labeling~\citep{huang2026labeling}. Marginal SCER is instead
the ratio of expected accepted errors to expected acceptance. Selective
conformal risk control studies selected prediction sets~\citep{xu2025selective}.
These distinctions concern estimands and decision rules, rather than a
general ordering of guarantee strength.

\paragraph{Risk calibration and algorithmic stability.}
Risk-controlling prediction sets accommodate losses beyond set
miscoverage~\citep{bates2021distribution}. LEC expresses marginal
selection-conditioned risk as a linear expectation
constraint~\citep{wang2026lec}. For adaptive non-monotonic calibration,
the full-data/leave-one-out analysis of \citet{angelopoulos2026conformal}
connects a reference rule to a calibration-fitted rule, building on the
broader relationship between stability and
generalization~\citep{bousquet2002stability}. Our analysis accounts for the reserve and signed stability gap,
and specializes the crossing bounds to anchored threshold search.
The explicit bounds certify a nonpositive linear-risk endpoint without
estimating stability by a bootstrap plug-in.

\begin{table}[t]
\centering
\small
\begin{tabular}{lrrrr}
\toprule
\rowcolor{SCHead}
Dataset & Cells & Higher & Equal & Lower\\
\midrule
CommonsenseQA & 20 & 20 & 0 & 0\\
ARC-Challenge & 20 & 14 & 0 & 6\\
TriviaQA & 20 & 17 & 3 & 0\\
\midrule
All & 60 & 51 & 3 & 6\\
\bottomrule
\end{tabular}
\caption{Retention relative to margin/KL SC-D over all primary cells
and 100 paired repetitions. LTT-cal and LTT-pool give the same counts.
All three ties have zero acceptance and undefined SCER. LTT-cal uses
$n$ labels. LTT-pool and SC-D use $m+n$ in different roles, with different
guarantee objects. Counts summarize observed retention differences across configurations.}
\label{tab:ltt-primary-summary}
\end{table}

\paragraph{Comparison with Learn then Test.}
Learn then Test (LTT) combines valid tests with multiplicity control
for high-probability risk guarantees, including non-monotonic
risks~\citep{angelopoulos2025learn}. Our LTT-CP instantiation tests
an independent selection-score grid with exact binomial tests and
Bonferroni correction. For a candidate $j$, let $N_j$ and $K_j$ be its
accepted and erroneous-accepted counts on the testing sample. Its
one-sided test uses
\begin{equation}
 p_j=\Pr\{\operatorname{Bin}(N_j,\alpha)\le K_j\}.
 \label{eq:ltt-selective-test}
\end{equation}
Under i.i.d. sampling, conditional on $N_j$, the error count is
binomial with parameter equal to that fixed candidate's selective risk.
The predictor, score, and all candidates must be independent of the
testing observations. We test all $L=20$ candidates at $\delta/L$ and
select the largest passing threshold; candidates with $N_j=0$ cannot
pass, and no passing candidate yields abstention. This is equivalent to
requiring a one-sided Clopper--Pearson upper bound at level
$1-\delta/L$ to be at most $\alpha$. Bonferroni allows dependence
between the candidate tests and protects the subsequent selection.
Thus LTT certifies the realized fitted policy, whereas SC-LEC's upstream
event certifies calibration-averaged marginal risk for a scan over
final-calibration order statistics. LTT may be preferable for a fixed
deployment, and valid testing constructions can also accommodate
adaptive candidates. The present experiments do not establish
an operational advantage over LTT at a common guarantee.

The complete primary comparison covers twelve dataset/model pairs
and five prespecified targets. Table~\ref{tab:ltt-primary-summary}
summarizes retention relative to SC-D. Both LTT variants are higher
in 51 of 60 cells, equal in three, and lower in six. On CommonsenseQA
Qwen2.5-32B at $\alpha=0.20$, pooled LTT retains
99.91\% with 15.98\% empirical SCER, versus SC-D's 77.52\% and
7.76\%. Both use $m+n=6572$ fitting labels, but LTT tests the pooled
labels while SC-D separates certification and calibration. LTT-cal
instead uses $n$ labels. All methods share predictions and paired test roles. Tables~\ref{tab:primary-csqa}--\ref{tab:primary-trivia}
report all 60 primary cells, including abstentions and undefined SCER cases. Availability denotes a
non-abstaining fitted policy, rather than the probability that a population
guarantee holds.

\clearpage
\begin{table*}[p]
\centering
\begingroup
\footnotesize
\begin{tabular}{lcrrrrrr}
\toprule
\rowcolor{SCHead}
Model & $\alpha$ & LEC & Anchor & SC-S & SC-D & LTT-cal & LTT-pool\\
\midrule
\multirow{5}{*}{Qwen2.5-7B} & 0.05 & 56.33 / 5.01 & 11.45 / 1.91 & 0.00 / -- & 11.45 / 1.91 & 44.01 / 3.06 & 48.49 / 3.60\\
 & 0.10 & 73.59 / 10.00 & 47.44 / 3.48 & 0.49 / 3.56 & 49.28 / 3.73 & 64.72 / 7.31 & 66.96 / 7.99\\
 & 0.15 & 87.54 / 14.89 & 54.83 / 4.74 & 56.81 / 5.15 & 59.58 / 5.85 & 79.64 / 12.12 & 81.49 / 12.75\\
 & 0.20 & 99.77 / 19.43 & 58.88 / 5.59 & 64.06 / 7.09 & 67.23 / 8.03 & 93.02 / 16.88 & 94.90 / 17.50\\
 & 0.30 & 99.96 / 19.53 & 69.35 / 8.76 & 83.68 / 13.56 & 87.58 / 14.96 & 99.90 / 19.49 & 99.90 / 19.49\\
\midrule
\multirow{5}{*}{Qwen2.5-32B} & 0.05 & 65.75 / 4.98 & 40.12 / 1.63 & 0.00 / -- & 40.18 / 1.63 & 55.17 / 3.18 & 58.60 / 3.55\\
 & 0.10 & 84.61 / 9.89 & 55.40 / 3.20 & 51.30 / 3.27 & 57.97 / 3.56 & 76.81 / 7.55 & 79.34 / 8.12\\
 & 0.15 & 97.50 / 14.95 & 60.52 / 3.96 & 63.79 / 4.60 & 66.42 / 5.15 & 90.21 / 11.90 & 91.82 / 12.56\\
 & 0.20 & 99.96 / 16.00 & 67.25 / 5.38 & 74.46 / 7.00 & 77.52 / 7.76 & 99.91 / 15.98 & 99.91 / 15.98\\
 & 0.30 & 99.96 / 16.00 & 79.36 / 8.15 & 97.64 / 15.00 & 99.75 / 15.91 & 99.91 / 15.98 & 99.91 / 15.98\\
\midrule
\multirow{5}{*}{Qwen3-8B} & 0.05 & 43.99 / 5.09 & 0.00 / -- & 0.00 / -- & 0.00 / -- & 18.02 / 3.41 & 32.48 / 3.75\\
 & 0.10 & 65.09 / 10.04 & 40.92 / 4.58 & 0.00 / -- & 41.44 / 4.65 & 55.11 / 7.30 & 58.31 / 8.03\\
 & 0.15 & 80.82 / 15.04 & 47.98 / 5.72 & 44.62 / 5.87 & 51.19 / 6.41 & 71.65 / 12.09 & 74.24 / 12.90\\
 & 0.20 & 94.67 / 20.04 & 54.30 / 7.12 & 57.45 / 7.93 & 60.73 / 8.77 & 85.87 / 16.83 & 88.18 / 17.74\\
 & 0.30 & 99.97 / 21.93 & 65.57 / 10.20 & 76.96 / 13.80 & 81.09 / 15.19 & 99.91 / 21.90 & 99.91 / 21.90\\
\midrule
\multirow{5}{*}{Qwen3-32B} & 0.05 & 55.60 / 4.91 & 7.14 / 2.02 & 0.00 / -- & 7.14 / 2.02 & 42.42 / 3.04 & 47.49 / 3.68\\
 & 0.10 & 76.26 / 10.04 & 47.76 / 3.77 & 0.00 / -- & 49.44 / 3.98 & 66.89 / 7.52 & 69.85 / 8.25\\
 & 0.15 & 90.32 / 14.96 & 54.80 / 4.81 & 56.65 / 5.16 & 59.39 / 5.74 & 82.50 / 12.02 & 84.59 / 12.77\\
 & 0.20 & 99.94 / 18.85 & 62.27 / 6.43 & 67.47 / 7.64 & 70.64 / 8.49 & 94.98 / 16.68 & 96.77 / 17.47\\
 & 0.30 & 99.96 / 18.87 & 73.26 / 9.28 & 88.33 / 14.22 & 92.34 / 15.74 & 99.89 / 18.82 & 99.89 / 18.82\\
\bottomrule
\end{tabular}

\endgroup
\caption{Complete CommonsenseQA primary comparison: retention / empirical SCER (\%) at five target risks, aggregated over all 100 paired repetitions. Anchor and SC methods use the prespecified margin/KL design. Failed and unavailable designs remain included; a dash denotes undefined SCER at zero acceptance. LTT-cal and LTT-pool use different fitting-label budgets, as described in the text.}
\label{tab:primary-csqa}
\end{table*}

\begin{table*}[p]
\centering
\begingroup
\footnotesize
\begin{tabular}{lcrrrrrr}
\toprule
\rowcolor{SCHead}
Model & $\alpha$ & LEC & Anchor & SC-S & SC-D & LTT-cal & LTT-pool\\
\midrule
\multirow{5}{*}{Qwen2.5-7B} & 0.05 & 84.81 / 5.08 & 19.88 / 2.23 & 0.00 / -- & 19.88 / 2.23 & 66.87 / 2.36 & 74.37 / 2.90\\
 & 0.10 & 98.11 / 9.79 & 72.06 / 2.69 & 0.74 / 3.50 & 74.95 / 3.09 & 87.60 / 5.93 & 91.34 / 7.20\\
 & 0.15 & 99.83 / 10.42 & 75.26 / 3.13 & 77.31 / 3.53 & 82.44 / 4.63 & 98.55 / 9.93 & 99.58 / 10.34\\
 & 0.20 & 99.83 / 10.42 & 80.56 / 4.13 & 87.13 / 5.95 & 92.84 / 7.83 & 99.58 / 10.34 & 99.58 / 10.34\\
 & 0.30 & 99.83 / 10.42 & 85.71 / 5.30 & 99.80 / 10.42 & 99.83 / 10.42 & 99.58 / 10.34 & 99.58 / 10.34\\
\midrule
\multirow{5}{*}{Qwen2.5-32B} & 0.05 & 99.55 / 4.51 & 77.47 / 0.78 & 0.00 / -- & 78.89 / 0.88 & 91.94 / 2.38 & 94.75 / 2.76\\
 & 0.10 & 99.86 / 4.68 & 85.05 / 1.39 & 87.21 / 1.66 & 91.14 / 2.21 & 99.63 / 4.55 & 99.63 / 4.55\\
 & 0.15 & 99.86 / 4.68 & 90.33 / 2.10 & 96.67 / 3.48 & 99.12 / 4.37 & 99.63 / 4.55 & 99.63 / 4.55\\
 & 0.20 & 99.86 / 4.68 & 93.29 / 2.56 & 99.83 / 4.67 & 99.86 / 4.68 & 99.63 / 4.55 & 99.63 / 4.55\\
 & 0.30 & 99.86 / 4.68 & 95.13 / 2.87 & 99.86 / 4.68 & 99.86 / 4.68 & 99.63 / 4.55 & 99.63 / 4.55\\
\midrule
\multirow{5}{*}{Qwen3-8B} & 0.05 & 89.66 / 5.14 & 57.98 / 1.41 & 0.00 / -- & 58.03 / 1.41 & 74.03 / 2.25 & 81.96 / 3.13\\
 & 0.10 & 99.53 / 9.15 & 74.05 / 2.29 & 13.28 / 3.16 & 77.61 / 2.76 & 92.44 / 6.14 & 95.00 / 6.98\\
 & 0.15 & 99.84 / 9.31 & 81.84 / 3.21 & 85.23 / 3.97 & 90.09 / 5.32 & 99.53 / 9.14 & 99.62 / 9.17\\
 & 0.20 & 99.84 / 9.31 & 84.50 / 3.63 & 92.73 / 6.20 & 97.72 / 8.25 & 99.62 / 9.17 & 99.62 / 9.17\\
 & 0.30 & 99.84 / 9.31 & 87.48 / 4.44 & 99.84 / 9.31 & 99.84 / 9.31 & 99.62 / 9.17 & 99.62 / 9.17\\
\midrule
\multirow{5}{*}{Qwen3-32B} & 0.05 & 99.48 / 4.61 & 81.07 / 0.85 & 0.00 / -- & 82.60 / 0.98 & 91.42 / 2.09 & 94.36 / 2.79\\
 & 0.10 & 99.84 / 4.73 & 86.01 / 1.34 & 88.37 / 1.64 & 92.06 / 2.33 & 99.59 / 4.62 & 99.61 / 4.63\\
 & 0.15 & 99.84 / 4.73 & 89.55 / 1.76 & 96.51 / 3.52 & 99.07 / 4.45 & 99.61 / 4.63 & 99.61 / 4.63\\
 & 0.20 & 99.84 / 4.73 & 91.27 / 2.00 & 99.81 / 4.72 & 99.84 / 4.73 & 99.61 / 4.63 & 99.61 / 4.63\\
 & 0.30 & 99.84 / 4.73 & 94.10 / 2.74 & 99.84 / 4.73 & 99.84 / 4.73 & 99.61 / 4.63 & 99.61 / 4.63\\
\bottomrule
\end{tabular}

\endgroup
\caption{Complete ARC-Challenge primary comparison: retention / empirical SCER (\%) at five target risks, aggregated over all 100 paired repetitions. Anchor and SC methods use the prespecified margin/KL design. Failed and unavailable designs remain included; a dash denotes undefined SCER at zero acceptance. LTT-cal and LTT-pool use different fitting-label budgets, as described in the text.}
\label{tab:primary-arc}
\end{table*}

\begin{table*}[p]
\centering
\begingroup
\footnotesize
\begin{tabular}{lcrrrrrr}
\toprule
\rowcolor{SCHead}
Model & $\alpha$ & LEC & Anchor & SC-S & SC-D & LTT-cal & LTT-pool\\
\midrule
\multirow{5}{*}{Qwen2.5-7B} & 0.05 & 2.58 / 7.33 & 0.00 / -- & 0.00 / -- & 0.00 / -- & 0.00 / -- & 0.00 / --\\
 & 0.10 & 17.85 / 9.97 & 0.00 / -- & 0.00 / -- & 0.00 / -- & 0.52 / 7.46 & 3.08 / 8.03\\
 & 0.15 & 26.99 / 14.86 & 0.00 / -- & 0.00 / -- & 0.00 / -- & 15.86 / 9.93 & 20.08 / 10.76\\
 & 0.20 & 34.69 / 19.89 & 0.40 / 12.19 & 0.00 / -- & 0.40 / 12.19 & 25.61 / 14.11 & 27.93 / 15.64\\
 & 0.30 & 48.71 / 29.85 & 25.05 / 13.68 & 0.00 / -- & 26.64 / 14.74 & 41.10 / 24.40 & 42.82 / 25.59\\
\midrule
\multirow{5}{*}{Qwen2.5-32B} & 0.05 & 8.43 / 6.25 & 0.00 / -- & 0.00 / -- & 0.00 / -- & 0.00 / -- & 0.00 / --\\
 & 0.10 & 38.68 / 10.18 & 0.00 / -- & 0.00 / -- & 0.00 / -- & 19.12 / 7.09 & 29.17 / 7.60\\
 & 0.15 & 54.10 / 15.24 & 24.30 / 8.09 & 0.00 / -- & 24.34 / 8.10 & 43.83 / 11.58 & 46.46 / 12.27\\
 & 0.20 & 63.79 / 20.23 & 37.15 / 9.86 & 0.00 / -- & 38.96 / 10.35 & 55.92 / 16.10 & 57.66 / 16.93\\
 & 0.30 & 83.27 / 29.92 & 47.73 / 12.85 & 51.41 / 14.26 & 55.99 / 16.19 & 75.04 / 25.86 & 77.13 / 26.88\\
\midrule
\multirow{5}{*}{Qwen3-8B} & 0.05 & 4.85 / 6.51 & 0.00 / -- & 0.00 / -- & 0.00 / -- & 0.00 / -- & 0.00 / --\\
 & 0.10 & 21.38 / 10.28 & 0.00 / -- & 0.00 / -- & 0.00 / -- & 2.87 / 7.42 & 11.46 / 7.08\\
 & 0.15 & 29.42 / 15.13 & 0.00 / -- & 0.00 / -- & 0.00 / -- & 20.52 / 9.96 & 23.52 / 11.49\\
 & 0.20 & 37.04 / 20.18 & 5.78 / 10.48 & 0.00 / -- & 5.78 / 10.48 & 28.61 / 14.68 & 30.36 / 15.78\\
 & 0.30 & 50.96 / 30.09 & 26.26 / 13.18 & 0.00 / -- & 28.71 / 14.70 & 43.30 / 24.55 & 45.27 / 26.00\\
\midrule
\multirow{5}{*}{Qwen3-32B} & 0.05 & 14.47 / 5.19 & 0.00 / -- & 0.00 / -- & 0.00 / -- & 0.10 / 2.97 & 0.39 / 4.75\\
 & 0.10 & 26.43 / 10.31 & 0.00 / -- & 0.00 / -- & 0.00 / -- & 14.67 / 5.61 & 18.83 / 6.96\\
 & 0.15 & 33.48 / 15.05 & 0.00 / -- & 0.00 / -- & 0.00 / -- & 26.54 / 10.47 & 28.35 / 11.40\\
 & 0.20 & 39.21 / 19.92 & 23.19 / 9.10 & 0.00 / -- & 23.26 / 9.14 & 32.75 / 14.58 & 34.42 / 15.79\\
 & 0.30 & 51.66 / 29.87 & 28.96 / 11.84 & 0.60 / 12.84 & 32.79 / 14.60 & 43.82 / 24.07 & 45.77 / 25.65\\
\bottomrule
\end{tabular}

\endgroup
\caption{Complete TriviaQA primary comparison: retention / empirical SCER (\%) at five target risks, aggregated over all 100 paired repetitions. Anchor and SC methods use the prespecified margin/KL design. Failed and unavailable designs remain included; a dash denotes undefined SCER at zero acceptance. LTT-cal and LTT-pool use different fitting-label budgets, as described in the text.}
\label{tab:primary-trivia}
\end{table*}

\end{document}